\documentclass[10pt,journal,compsoc,twocolumn ]{IEEEtran}

\usepackage{amsfonts}
\usepackage{mathrsfs}
\usepackage{graphicx}
\usepackage{amsmath}
\usepackage{amssymb}
\usepackage{multicol}
\usepackage{multirow}
\usepackage{threeparttable}

\usepackage{epstopdf}

\usepackage{times}
\usepackage{epsfig}
\usepackage{bbm}
\usepackage{rotating}
\usepackage{array}
\allowdisplaybreaks[4]

\usepackage{subfigure}
\usepackage{bm}
\usepackage{algorithm}
\usepackage{algorithmic}
\begin{document}
% first the title is needed

\title{Cycle Label-Consistent Networks for Unsupervised Domain Adaptation}

\author{Mei Wang, Weihong Deng
\thanks{Mei Wang and Weihong Deng are with the Pattern Recognition and Intelligent System Laboratory, School of Artificial Intelligence, Beijing University of Posts and Telecommunications, Beijing, 100876, China. E-mail: \{wangmei1,whdeng\}@bupt.edu.cn. (Corresponding Author: Weihong Deng)}}
\maketitle

%%%%%%%%% ABSTRACT
\begin{abstract}

Domain adaptation aims to leverage a labeled source domain to learn a classifier for the unlabeled target domain with a different distribution. Previous methods mostly match the distribution between two domains by global or class alignment. However, global alignment methods cannot achieve a fine-grained class-to-class overlap; class alignment methods supervised by pseudo-labels cannot guarantee their reliability. In this paper, we propose a simple yet efficient domain adaptation method, i.e. Cycle Label-Consistent Network (CLCN), by exploiting the cycle consistency of classification label, which applies dual cross-domain nearest centroid classification procedures to generate a reliable self-supervised signal for the discrimination in the target domain. The cycle label-consistent loss reinforces the consistency between ground-truth labels and pseudo-labels of source samples leading to statistically similar latent representations between source and target domains. This new loss can easily be added to any existing classification network with almost no computational overhead. We demonstrate the effectiveness of our approach on MNIST-USPS-SVHN, Office-31, Office-Home and Image CLEF-DA benchmarks. Results validate that the proposed method can alleviate the negative influence of falsely-labeled samples and learn more discriminative features, leading to the absolute improvement over source-only model by 9.4\% on Office-31 and 6.3\% on Image CLEF-DA.
\end{abstract}


%%%%%%%%% BODY TEXT
\section{Introduction}

Deep learning methods have propelled unprecedented advances in a wide variety of visual recognition tasks, such as image recognition \cite{krizhevsky2012imagenet} and object detection \cite{ren2015faster}, demonstrating an excellent generalization ability. The recent success of deep learning heavily depends on large quantities of labeled data. However, collecting and annotating datasets for every new task and domain are extremely expensive and time-consuming processes, sufficient training data may not always be available. Training deep models on available labeled data from different but related source domains is a strong motivation to reduce the labeling consumption. But, analogously to other statistical machine learning techniques, this learning paradigm suffers from the domain shift problem \cite{donahue2014decaf,torralba2011unbiased}, which degrades the performance of pre-trained models when testing on target domains. In this regard, the research community has proposed different mechanisms, unsupervised domain adaptation (UDA) \cite{wang2018deep} is one of the promising methodologies to address this problem, which learns a good predictive model for the target domain using labeled examples from the source domain but only unlabeled examples from the target domain.

\begin{figure}[htbp]
\centering
\includegraphics[width=8cm]{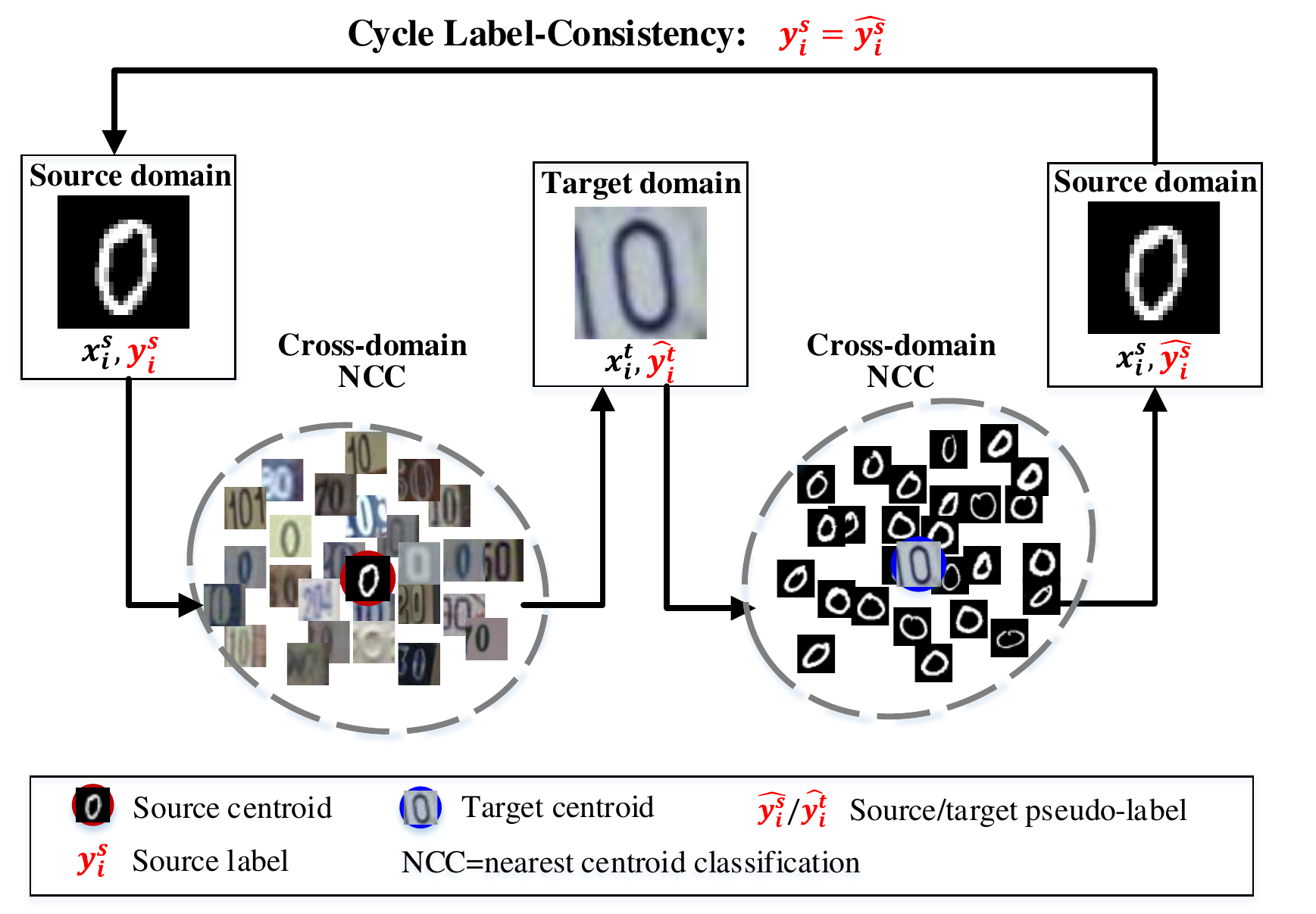}
\caption{ Illustration of the main idea in this work. If we classify the target samples according to the ground-truth labels of source samples, and then classify the source samples according to pseudo-labels of target samples, we should assign source samples back to the original labels.}
\label{fig0}
\end{figure}

Many deep UDA methods \cite{wang2018deep} try to match global distributions of source and target data to learn domain-invariant features, and then directly apply the classifier learned from only source labels to target domain \cite{Tzeng2014Deep,Ganin2015Unsupervised,courty2017optimal,zellinger2017central}. Other studies \cite{chen2018progressive,hou2016unsupervised,li2018domain} recently propose to utilize pseudo-labels to take category information into consideration and learn target discriminative representations. %However, matching the discrepancy at the domain level usually neglects the class from which the samples are drawn. Therefore, there is no guarantee that samples from different domains but with the same class label will map nearby in the feature space, the latent space may not form a single compact cluster per class so that the learned decision boundary may generalize poorly.
These target pseudo-labels are generated by maximum posterior probability of source softmax classifier, and then are utilized by following two approaches. 1) These pseudo labels are directly used to finetune the network by supervised losses \cite{saito2017asymmetric,zhang2018collaborative}. However, due to large domain shift, the learned source classifiers might be incapable of precisely labeling target samples with an expected accuracy requirement. Supervising with these target pseudo-labels may lead to the error accumulation caused by some falsely pseudo-labeled samples. 2) The generated pseudo labels are used to align the centroids of source and target samples belonged to the same category \cite{xie2018learning,chen2018progressive}. Although the target centroids computed by the mean vectors of embeddings of their constituent samples can alleviate the error generated by some mislabelled samples, these centroids correspondingly miss some backpropagation information provided by samples.

In this paper, %inspired of Cycle-Consistent Adversarial Networks (CycleGAN) \cite{zhu2017unpaired} and Prototypical Networks \cite{snell2017prototypical},
we propose a simple yet efficient domain adaptation method, i.e. Cycle Label-Consistent Network (CLCN), exploiting cycle label-consistency and cross-domain nearest centroid classification (NCC) algorithm to learn aligned and discriminative presentations for the target domain, as shown in Fig. \ref{fig0}. Different from other domain adaptation methods, CLCN is supervised with ground-truth labels and optimizes networks based on backpropagation information provided by each sample. During training, in addition to minimizing the cross-entropy loss on labeled source data, we reinforces the consistency between ground-truth labels and pseudo-labels of source samples leading to statistically similar latent representations between source and target domains. First, CLCN proposes to utilize cross-domain nearest centroid classification algorithm to estimate pseudo-labels for target domain and source domain in turn. The centroid of each source class is computed by the mean vector of embeddings of its constituent instances. With the help of source centroids, target pseudo-labels can be obtained by cross-domain classification which is performed for each target point by simply finding the nearest source centroids. Then, we update the target centroids in the same way as source centroids, and assign each source sample to its closest target centroid to get source pseudo-label. Such cross-domain classification enables to transport label information from the source domain to the target domain and back to the source domain. After obtaining source pseudo-labels, we reinforce the consistency between pseudo-labels and ground-truth labels of source samples by computing cross entropy between them so that statistically similar latent representations between source and target domains are learned at class level. Intuitively, with similar latent representations, the target pseudo-labels can be more accurately inferred from the labeled source samples.

Compared with other methods which directly use target pseudo-labels to finetune network, our CLCN just utilizes them to compute the target centroids and transport label information to source domain. The centroids computed by averaging the features of each class can alleviate the wrong information of several falsely-labeled samples. And cycle label-consistent loss is supervised with ground-truth labels of source samples, which makes our CLCN be more reliable and take full advantage of backpropagation information provided by each sample. Moreover, our source pseudo-labels here are ``soft" (label probability) which are produced by computing a distribution over target pseudo-centroids for each source sample. Cycle consistency computed by ``soft" pseudo-labels imposes a stricter restriction and obtains more compact clusters in latent space.

Following the standard evaluation protocol in the unsupervised domain adaptation community, we evaluate our method on the digit classification task using MNIST \cite{lecun1998gradient}, SVHN \cite{netzer2011reading} and USPS \cite{denker1989neural} as well as the object recognition task using the Office-31 \cite{saenko2010adapting}, Office-Home \cite{venkateswara2017deep} and ImageCLEF-DA dataset \cite{caputo2014imageclef}, and demonstrate the superiority of the proposed method. To summarize, the contributions of the paper are threefold:
\begin{itemize}
  \item First, we propose a novel domain adaptation method, i.e. Cycle Label-Consistent Networks (CLCN), by exploring an intrinsic property that classification should be ¡°cycle consistent¡±, in the sense that if we classify the target samples according to the ground-truth labels of source samples, and then classify the source samples according to pseudo-labels of target samples, we should assign source samples back to the original labels.
  \item Second, we formulate a cross-domain nearest centroid classification (NCC) algorithm to transport label information from the source domain to the target domain and back to the source domain. Especially, the label probability is computed by the cross-domain sample-to-centroid distances, and these ``soft" pseudo-labels encourage stricter restriction and more compact clusters compared with ``hard" pseudo-labels.
  \item Third, benefiting from being supervised with ground-truth labels, our CLCN can alleviate the negative influence of falsely-labeled samples and take full advantage of backpropagation information provided by each sample. We experimentally show that CLCN can achieve comparable performance with other complicated domain adaptation methods only by a source classification loss and a simple penalty item.
\end{itemize}

The rest of this paper is organized as follows. In the next section, we briefly review related work on deep unsupervised domain adaptation and cycle consistency. Section III describes the proposed method including cross-domain nearest centroid classification algorithm and cycle label-consistent loss. Additionally, experimental results are shown and analyzed in Section IV. Section V offers the concluding remarks.

\section{Related work}

\subsection{Unsupervised domain adaptation}

There is always a distribution change or domain gap between training and testing sets, which would degrade the performance. Yosinski et al. \cite{yosinski2014transferable} comprehensively explored the transferability of deep neural networks and finetuned the network with sufficient target labeled data to improve performance on target domain. However, in practical scenario, labeled target data is usually limited or unacquirable. To address this issue, many UDA approaches \cite{wang2018deep} are proposed, as shown in Table \ref{referen}.

\begin{table*}[htbp]
	\begin{center}
	\caption{Comparison of existing unsupervised domain adaptation methods. The average accuracies tested on Office-31 dataset are reported.}
    \setlength{\tabcolsep}{2mm}{
	\begin{tabular}{l|lll}
        \hline
        % after \\: \hline or \cline{col1-col2} \cline{col3-col4} ...
         Category & Algorithm &  Training method & Accuracy \tnote{1}\\ \hline\hline
          & DDC \cite{Tzeng2014Deep}, DAN \cite{Long2015Learning} & MMD & 72.9 (Alexnet)\\
           & JAN \cite{long2017deep} & JMMD & 76.0 (Alexnet)\\
          & RTN \cite{long2016unsupervised} & MMD, residual learning & 73.7 (Alexnet) \\ 
          & WMMD \cite{yan2017mind} & weighted MMD & 72.1 (Alexnet)\\ 
          Global  & CMD \cite{zellinger2017central} & CMD & 79.9 (Alexnet) \\
          alignment & Deep CORAL \cite{sun2016deep} & CORAL & 72.1 (Alexnet)\\ 
          & DANN \cite{Ganin2015Unsupervised} & adversarial learning & 74.3 (Alexnet) \\
          & CAADA \cite{rahman2019correlation} & adversarial learning, CORAL & 78.3 (Alexnet)\\
          & SymNet \cite{zhang2019domain} & two-level adversarial learning & 88.4 (Resnet50)\\
          & Cicek et al. \cite{cicek2019unsupervised} & 2K-way adversarial learning & - \\
          & STA \cite{liu2019separate}, UAN \cite{you2019universal} & weighted adversarial learning & 89.2 (Resnet50)\\ 
          & SAFN \cite{xu2019larger} & feature-norm alignment & 87.6 (Resnet50)\\ \hline
         Local & OT \cite{courty2017optimal}, SWD \cite{lee2019sliced} & optimal transport &- \\
         alignment & Das et al. \cite{das2018sample,das2018graph,das2018unsupervised} & graph matching& -\\ \hline
         & AsmTri \cite{saito2017asymmetric} & tri-training, finetuning & -\\
         & MSTN \cite{xie2018learning} & class alignment & 79.1 (Alexnet) \\
         Pseudo & PFAN \cite{chen2018progressive} & class alignment, progressive learning & 80.4 (Alexnet)\\
         label & iCAN \cite{zhang2018collaborative} & progressive learning, adversarial learning & 87.2 (Resnet50)\\
         & PACET \cite{liang2019exploring} & progressive learning, weighted targets & 76.6 (Alexnet)\\
         & GCAN \cite{ma2019gcan} & structure-,domain-, class-alignment &  80.6 (Alexnet)\\\hline
        \hline
	\end{tabular}}
    \end{center}
    \label{referen}
\end{table*}

\textbf{Global-alignment based method}. Some papers explore domain-invariant feature spaces by minimizing global discrepancy measured by statistic loss \cite{Tzeng2014Deep,Long2015Learning,sun2016deep} and adversarial loss \cite{Tzeng2017Adversarial,Ganin2015Unsupervised,tzeng2015simultaneous}. Maximum mean discrepancy (MMD) \cite{Tzeng2014Deep,Long2015Learning,long2016unsupervised,yan2017mind}, Central Moment Discrepancy (CMD) \cite{zellinger2017central} and Correlation alignment (CORAL) \cite{sun2016deep} are commonly-used statistic losses for UDA. For example, in Deep Domain Confusion (DDC) \cite{Tzeng2014Deep}, the network is optimized by classification loss in the source domain, while domain difference is minimized by one adaptation layer with the MMD metric. % Long et al. \cite{Long2015Learning} proposed Deep Adaptation Networks (DAN) that matches the shift in marginal distributions across domains by adding multiple adaptation layers and exploring multiple kernels. %Zong et al. \cite{zong2018domain} proposed to regenerate the source and target samples with the MMD regularization term such that the regenerated source and target micro-expression samples would share the same or similar feature distributions. %A joint adaptation network (JAN) \cite{Long2016Deep} aligned the shift in the joint distributions of input features and output labels based on a joint maximum mean discrepancy (JMMD) criterion. %More recently, Yan et al. \cite{Yan2017Mind} proposed a weighted MMD model that introduces an auxiliary weight for each class in the source domain when the class weights in the target domain are not the same as those in the source domain.
%Inspired by the recent success of generative adversarial networks (GAN) \cite{goodfellow2014generative},
Adversarial learning has shown great promise for use in domain adaptation. %Adversarial loss makes the distribution of both domains similar enough through domain classifier such that the network is fooled and can be directly used in the target domain. 
The domain-adversarial neural network (DANN) \cite{Ganin2015Unsupervised} integrated a gradient reversal layer (GRL) to train a feature extractor by maximizing the domain classifier loss and simultaneously minimizing the label predictor loss. Rahman et al. \cite{rahman2019correlation} proposed a correlation-aware adversarial domain adaptation framework where the correlation metric is used jointly with adversarial learning to minimize the domain disparity of the source and target data. Domain-symmetric networks (SymNets) overcomes the limitation in aligning the joint distributions of feature and category across domains via two-level domain confusion losses. Separate to Adapt (STA) \cite{liu2019separate} and Universal Adaptation Network (UAN) \cite{you2019universal} use adversarial learning to align the shared classes of target domain and source domain, and reject samples of unknown classes to address open set DA problem.
%Other methods typically learn the domain-invariant representation by a reconstruction loss in the source and target domains. The domain separation networks (DSNs) \cite{bousmalis2016domain} found the shared representation by reconstructing the input sample with a shared representation that is similar for both domains and a private representation that is domain specific. In \cite{ghifary2016deep}, Grifary et al. proposed the deep reconstruction classification network (DRCN) method to learn a shared representation by supervised classification of labeled source data and unsupervised reconstruction of unlabeled target data.

\textbf{Local-alignment based method}. These methods \cite {das2018sample,das2018graph} place an emphasis on establishing a sample-to-sample correspondence between each source sample and each target sample to mitigate domain gap. Compared with global-alignment based methods, they consider the effect of each and every sample in the dataset explicitly. For example, Courty et al. \cite{courty2017optimal} learned a transport plan for each source sample so that they are close to the target samples. Their transport plan is defined on a first-order, point-wise unary cost between each source sample and each target sample. Das et al. \cite{das2018sample,das2018graph,das2018unsupervised} developed a framework that exploit all the first-, second- and third-order relations to match the source and target samples along with a regularization using labels of the source data. Such higher order relations provide additional geometric and structural information about the data beyond the unary point-wise relations.

\textbf{Pseudo-label based method}. Target pseudo-labels \cite{chen2011co,liang2019exploring,chen2020deep} are utilized to finetune the network so that the lack of categorical information is compensated and discriminative representations are learned in the target domain. Saito et al. \cite{saito2017asymmetric} introduced the idea of tri-training \cite{zhou2005tri} into domain adaptation. Two different networks assign pseudo-labels to unlabeled samples through voting, another network is trained by these pseudo-labels to learn target discriminative representations. However, the generated pseudo-labels may be unreliable because of large domain shift. Supervising with these target pseudo-labels may lead to the error accumulation caused by falsely pseudo-labeled samples. Some methods propose to utilize progressive learning to tackle this error accumulation. Zhang et al. \cite{zhang2018collaborative} iteratively selected confident pseudo-labels based on the classifier from the previous training epoch and re-trained the model by using the enlarged training set. Liang et al. \cite{liang2019exploring} tackled the uncertainty in pseudo labels of the target domain from two aspects, progressive target instance selection and incorporating the learned class confidence scores to characterize both the within- and cross- domain relations. Other methods utilize pseudo labels to perform centroid alignment instead of finetuning to alleviate the negative effect of these falsely-labeled samples. Combining with adversarial loss, the moving semantic transfer network (MSTN) \cite{xie2018learning} took semantic information into consideration for unlabeled target samples by aligning labeled source centroids and pseudo-labeled target centroids. Chen et al. \cite{chen2018progressive} combined easy-to-hard transfer strategy and centroid alignment, resulting in an improved target classification accuracy. Graph Convolutional Adversarial Network (GCAN) \cite{ma2019gcan} utilized the concatenated CNN and GCN features to generate pseudo-labels and performed structure-aware alignment, domain alignment, and class centroid alignment to mitigate domain gap. However, these centroid alignment methods may correspondingly miss some backpropagation information provided by samples. In this paper, we exploit cycle label-consistency and cross-domain nearest centroid classification algorithm to address the error accumulation problem and take full advantage of backpropagation information, leading to improved performance on target domain.

\subsection{Cycle consistency}

The idea of using transitivity as a way to regularize structured data has a long history, for example, higher-order cycle consistency has been used in structure from motion \cite{zach2010disambiguating}, 3D shape matching \cite{huang2013consistent}, co-segmentation \cite{wang2013image}, dense semantic alignment \cite{zhou2015flowweb,zhou2016learning}, and depth estimation \cite{godard2017unsupervised}.

In domain adaptation field, cycle-consistency was first introduced into deep domain adaptation by Zhu et al. \cite{zhu2017unpaired}. Zhu et al. proposed Cycle-Consistent Adversarial Networks (CycleGAN) \cite{zhu2017unpaired} which can translate one image from source domain $X$ into target domain $Y$ in the absence of any paired training examples. CycleGAN learns a mapping $G:X\rightarrow Y$ and an inverse mapping $F:Y\rightarrow X$. Two discriminators measure how realistic the generated image is ($G(X)\approx Y$ or $G(Y)\approx X$) by an adversarial loss and how well the original input is reconstructed after a sequence of two generations ($F(G(X))\approx X$ or $G(F(Y))\approx Y$) by a cycle-consistent loss. The dual-GAN \cite{yi2017dualgan} and the disco-GAN \cite{kim2017learning} were proposed at the same time, whose core idea is similar to CycleGAN. Cycle-Consistent Adversarial Domain Adaptation (CyCADA) \cite{hoffman2017cycada} unifies prior feature-level and image-level adversarial domain adaptation methods together with cycle-consistent image-to-image translation techniques in which cycle-consistent loss encourages the cross-domain transformation to preserve local structural information. Associative domain adaptation \cite{haeusser2017associative} reinforces associations between source and target data directly in embedding space. It defines associative similarity as two-step transition probability of an imaginary random walker starting from an embedding of source domain and returning to another embedding via target domain embedding, and constraints the two-step probabilities to be similar to the uniform distribution over the class labels via a cross-entropy loss. Transduction with Domain Shift (TDS) \cite{sener2016learning} is most similar to our work, which combined metric learning and cycle consistency. It generates target pseudo-labels based on the k-nearest-neighbor rule, and the cycle consistency between ground-truth labels and pseudo-labels of source samples is enforced without explicitly computing source pseudo-labels but using the large-margin nearest neighbor (LMNN) \cite{weinberger2009distance} rule.

\section{Cycle label-consistent network}

\subsection{Problem formulation}

\begin{figure*}[htbp]
\centering
\includegraphics[width=14cm]{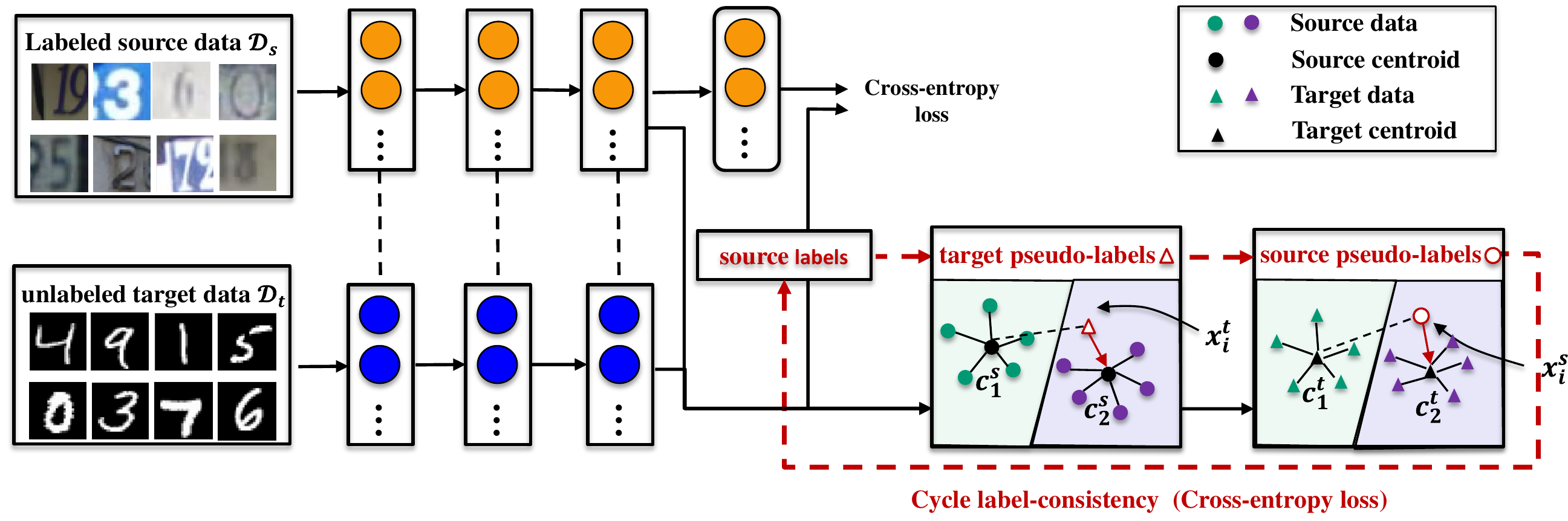}
\caption{ The overall structure of our CLCN method. The inputs of the upper network are source labeled images while those of the lower are target unlabeled data. In addition to minimizing the cross-entropy loss on labeled source data, cross-domain nearest centroid classification algorithm estimates target pseudo-labels through source centroids and estimates source pseudo-labels through target pseudo-centroids in turn. Then, CLCN reinforces the consistency between the ground-truth labels and the pseudo-labels of source samples resulting in statistically similar latent representations between source and target domains and compact clusters in latent space.}
\label{fig1}
\end{figure*}

\begin{table}[htbp]
\small
\caption{Summary of major notations used in the paper. }
	\begin{center}
    \setlength{\tabcolsep}{1mm}{
	\begin{tabular}{ll}
        \hline
        % after \\: \hline or \cline{col1-col2} \cline{col3-col4} ...
        Notations & Description \\ \hline
        $\mathcal{D}_{s}$/$\mathcal{D}_{t}$ & the source/target domain  \\
        $x^{s}_{i}$/$x^{t}_{i}$ &  sample in the source/target domain \\
        $N_s$/$N_t$ & number of source/target samples \\
        $y^{s}_{i}$ & true label of source sample $x^{s}_{i}$ \\
        $\hat{y^{s}_{i}}$/$\hat{y^{t}_{i}}$ & pseudo label of source/target sample $x^{s}_{i}$/$x^{t}_{i}$ \\
        $x^{s}_{i,k}$ & source data with true label $k$ \\
        $x^{s}_{i,\hat{k}}$/$x^{t}_{i,\hat{k}}$ & source/target data with pseudo label $\hat{k}$ \\
        $\mathcal{D}^{s}_k$ & the set of all source images with true label $k$ \\
        $\mathcal{D}^{s}_{\hat{k}}$/$\mathcal{D}^{t}_{\hat{k}}$ & the set of source/target images with pseudo-label $\hat{k}$ \\
        $c^{s}_{k}$ & centroid of source data with true label $k$ \\
        $c^{t}_{\hat{k}}$ & centoid of target data with pseudo label $\hat{k}$ \\
        $N_k^s$ & number of source samples with true label $k$ \\
        $N_{\hat{k}}^{s}$/ $N_{\hat{k}}^{t}$ & number of source/target samples with pseudo label $\hat{k}$ \\
        $K$ & number of source or target categories \\
        $\Theta$ & network parameters to be learned \\
        $\alpha,\theta$ & trade-off hyper-parameters \\
        \hline
	\end{tabular}}
    \end{center}
    \label{tab1}
\end{table}

In unsupervised domain adaptation, we are given a set of labeled data from the source domain, and denote them as $\mathcal{D}_{s}=\{{x^{s}_{i}},{y^{s}_{i}}\}^{N_s}_{i=1}$, where $x^{s}_{i}$ is the $i$-th source sample, $y^{s}_{i}$ is its category label, and $N_s$ is the number of source images. A set of unlabeled data from the target domain is given as well and is denoted as $\mathcal{D}_{t}=\{{x^{t}_{i}}\}^{N_t}_{i=1}$, where $x^{t}_{i}$ is the $i$-th target sample and $N_t$ is the number of target images. The data distributions of two domains are different, $P(X_s,Y_s)\neq P(X_t,Y_t)$. We assume that the source and target domains contain the same object classes. Our goal is to learn a classifier $h:\mathcal{X}\rightarrow \mathcal{Y}$ (parameterized by $\Theta$) that can decrease the domain discrepancy so as to minimize the target error by using the supervision information from source data. Generally, the classifier $h$ is constructed as $h=g\circ f$ where $f$ maps samples into features in the space $\mathcal{F}$ and $g$ outputs the predictions based on the extracted features. The learning process of our CLCN includes simultaneously optimizing the classifier $h$ w.r.t. the labeled source data, and reinforcing the consistency between pseudo-labels and ground-truth labels of source samples. For the convenience of reading, we list some of the major notations that are used throughout this paper in Table \ref{tab1}.

\subsection{Cycle label-consistent network}

In addition to minimizing the cross-entropy loss on labeled source data, our CLCN utilizes cross-domain nearest centroid classification algorithm to classify the target samples according to the ground-truth labels of source samples, and then classifies the source samples according to pseudo-labels of target samples, and finally reinforces consistency between ground-truth labels and pseudo-labels of source samples. By supervising with ground-truth labels, our cycle label-consistent loss alleviates the accumulation of errors and takes full advantage of backpropagation information provided by each sample, resulting in statistically similar latent representations between two domains and compact clusters in latent space. The overall architecture of CLCN is depicted in Fig. \ref{fig1}.

\textbf{Source2target nearest centroid classification.} This classification algorithm aims to estimate pseudo-labels in one domain with the help of centroids of classes in the other domain. First, for source domain, CLCN computes an $M$-dimensional representation $c^{s}_{k}$, or centroid, of $k$-th source class through the embedding function $f$ with learnable parameters $\Theta$. Each centroid is the mean vector of the embedded support points belonging to its class:
\begin{equation}
c_{k}^s=\frac{1}{N_k^s}\sum_{x_{i}^{s}\in \mathcal{D}^{s}_k}f(x_{i}^{s}) \label{center}
\end{equation}
where $\mathcal{D}^{s}_k$ denotes the set of all source images with label $k$ and $N_k^s$ denotes the number of source samples in $\mathcal{D}^{s}_k$. We assume the existence of an embedding space in which the projections of samples in each class cluster around a single centroid. Therefore, for each target sample $x^{t}_{i}$, we compute its cosine similarities with all source centroids, and assign it to corresponding centroid with the largest cosine similarity:
\begin{equation}
\begin{split}
&\hat{y_{i}^{t}}=arg \max\limits_{k}\ s_{k} \\
& {\rm where,} \ s_{k}=cos \left ( f(x_{i}^{t})  ,c_{k}^s\right ) \label{scattered}
\end{split}
\end{equation}
where $cos(\cdot,\cdot)$ denotes the cosine similarity function between two vectors. Finally, we can annotate each target node with pseudo label $\hat{y^{t}_{i}}$ and add them into the $\hat{k}$-th pseudo-class of target domain $\mathcal{D}^{t}_{\hat{k}}$.

\textbf{Target2source nearest centroid classification.} After generating target pseudo-labels with the help of source centroids, we additionally generate source pseudo-labels with the help of target pseudo-centroids. The centroid of $\hat{k}$-th target pseudo-class, i.e. $c^{t}_{\hat{k}}$, are computed in the same way:
\begin{equation}
c^{t}_{\hat{k}}=\frac{1}{N_{\hat{k}}^{t}}\sum_{x_{i}^{t}\in \mathcal{D}^{t}_{\hat{k}}}f(x_{i}^{t})
\end{equation}
where $\mathcal{D}^{t}_{\hat{k}}$ denotes the set of all target images with pseudo-labels $\hat{k}$ and $N_{\hat{k}}^t$ denotes the number of target samples in $\mathcal{D}^{t}_{\hat{k}}$. Then, the method of generating pseudo-labels for source domain is as follows. Given a source sample $x^{s}_{i}$, we compute the cross-domain sample-to-centroid distances between $x^{s}_{i}$ and all target pseudo-centroids, and directly produce its score distribution $p_{score}\left ( \hat{y}|x^{s}_{i} \right )\in \mathbb{R}^K$ over $K$ target classes via a softmax function on these cross-domain sample-to-centroid distances:
\begin{equation}
p_{score}\left ( \hat{y}=\hat{k}|x^{s}_{i} \right )=\frac{exp\left ( d\left ( f\left ( x^{s}_{i} \right ),c^{t}_{\hat{k}} \right ) \right )}{\sum_{j=1}^K exp\left ( d\left ( f\left ( x^{s}_{i} \right ),c^{t}_{j} \right ) \right )} \label{softmax}
\end{equation}
where $d(\cdot,\cdot)$ is the distance function between source sample and target centroid, and cosine distance is used in our paper; $\hat{k}$-th element of the score distribution, i.e. $p_s\left ( \hat{y}=\hat{k}|x^{s}_{i} \right )$, is the probability of $x^{s}_{i}$ belonging to pseudo-class $\hat{k}$. We treat the score distribution $p_{score}\left ( \hat{y}|x^{s}_{i} \right )$ as ``soft" pseudo-labels $\hat{y^{s}_{i}}$ of source samples.

Through such cross-domain nearest centroid classification algorithm, we finally transport label information from the source domain to the target domain and back to the source domain. Compared with generating pseudo-labels by softmax classifiers, our cross-domain nearest centroid classification algorithm can be performed easily with no extra structural and almost no training overhead, and can be more effective in cycle label-consistency, which has been proved in Section \ref{analysis}.

\textbf{Cycle label-consistency.} After classifying the target samples according to the ground-truth labels of source samples and then classifying the source samples according to pseudo-labels of target samples, we keep the consistency between predicted labels and ground-truth labels in source domain. Because such cycle label-consistency will be realized if the embedded features from both domains form well-aligned clusters. To establish this consistency, the model is optimized by minimizing the cross-entropy loss between pseudo-labels and ground-truth labels of source samples:
\begin{equation}
\mathcal{L}_{cyc}=-\frac{1}{N_s}\sum_{i=1}^{N_s} \sum_{\hat{k}=1}^{K}\mathbf{1}_[{\hat{k}=y^{s}_{i}}]log p_{score}\left ( \hat{y}|x^{s}_{i} \right ) \label{cycle}
\end{equation}
where $y^{s}_{i}$ and $p_{score}\left ( \hat{y}|x^{s}_{i} \right )$ are ground-truth labels and ``soft" pseudo-labels of source samples, respectively; $\mathbf{1}_[{\hat{k}=y^{s}_{i}}]$ is 1 when ${\hat{k}=y^{s}_{i}}$, otherwise, it is 0. Compared with ``hard" pseudo-labels of source samples which are generated in the same way as target pseudo-labels, minimizing $\mathcal{L}_{cyc}$ computed by ``soft" pseudo-labels imposes stricter constraint. It may encourage pseudo-labels and ground-truth labels to overlap rather than being nearest. Meanwhile, because ``soft" pseudo-label indicates cross-domain sample-to-centroid distances, such cycle label consistency also clusters each source sample into its centroid leading to compact cluster per class.

\subsection{Optimization}

We can incorporate the cycle label-consistent loss into any popular deep convolutional neural networks (CNNs) architecture (e.g., AlexNet, VGG, ResNet, DenseNet, etc.) to learn robust features for unsupervised domain adaptation. Let $p_c(y=k|x^{s}_{i})$ be the conditional probability that the CNN assigns $x^{s}_{i}$ to $k$-th class. The classification loss on labeled source data, i.e. cross-entropy of ground-truth labels and network predictions, can be denoted as:
\begin{equation}
\mathcal{L}_{C}=-\frac{1}{N_s}\sum_{i=1}^{N_s} \sum_{k=1}^{K}\mathbf{1}_[{k=y^{s}_{i}}]log p_c(y|x^{s}_{i}) \label{softmaxloss}
\end{equation}
where $y^{s}_{i}$ is ground-truth labels of source samples. $\mathbf{1}_[{k=y^{s}_{i}}]$ is 1 when ${k=y^{s}_{i}}$, otherwise, it is 0. In addition to minimizing the cross-entropy loss on labeled source data, we optimize cycle label-consistent loss to improve network performance on unlabeled target domain. Then, the final objective for our Cycle Label-Consistent Networks (CLCN) can be written as:
\begin{equation}
\mathcal{L}=\mathcal{L}_{C}+\alpha\mathcal{L}_{cyc}
\end{equation}
where $\alpha$ is trade-off hyper-parameter. The cycle label-consistent loss learns aligned and compact embeddings for the source and target samples, while the source classification loss minimizes the prediction error of the source data. The entire procedure of computing total loss of CLCN is depicted in Algorithm \ref{al1}.

\begin{algorithm}
\small
\caption{ Training episode loss computation for Cycle Label-Consistent Networks (CLCN).}
\label{al1}
\begin{algorithmic}[1]
\REQUIRE ~~\\
Source domain labeled samples $\mathcal{D}_{s}=\{{x^{s}_{i}},{y^{s}_{i}}\}^{N_s}_{i=1}$, and target domain unlabeled samples $\mathcal{D}_{t}=\{{x^{t}_{i}}\}^{N_t}_{i=1}$. Batch size $N$, the number of classes $K$, trade-off parameters $\alpha$ and $\theta$.
\ENSURE ~~\\
The loss $\mathcal{L}$ for a randomly generated training episode.
\STATE $\mathcal{S}$={\large R}ANDOM{\large S}AMPLE($\mathcal{D}_{s}$,$N$)
\STATE $\mathcal{T}$={\large R}ANDOM{\large S}AMPLE($\mathcal{D}_{t}$,$N$)
\STATE \textbf{\emph{Stage-1:}} // \textbf{source2target nearest centroid classification}
\STATE \textbf{for} k = 1 to K \textbf{do}
\STATE \ \ \ \ \ $c_{k(t)}^s \leftarrow \frac{1}{\left | \mathcal{S}_k \right |}\sum_{x_{i}^{s}\in \mathcal{S}_k}f(x_{i}^{s})$
\STATE \ \ \ \ \ $c_{k}^s \leftarrow \theta c_{k}^s+(1-\theta)c_{k(t)}^s$
\STATE \textbf{end for}
\STATE Assign target samples to its closest source centroids according to Eqn. \ref{scattered}, and obtain target pseudo-labels
\STATE \textbf{\emph{Stage-2:}} // \textbf{target2source nearest centroid classification}
\STATE \textbf{for} k = 1 to K \textbf{do}
\STATE \ \ \ \ \ $c^{t}_{\hat{k}} \leftarrow \frac{1}{\left | \mathcal{T}_{\hat{k}} \right |}\sum_{x_{i}^{t}\in \mathcal{T}_{\hat{k}}}f(x_{i}^{t})$
\STATE \ \ \ \ \ $c_{k}^t \leftarrow \theta c_{k}^t+(1-\theta)c_{k(t)}^t$
\STATE \textbf{end for}
\STATE For each source sample, compute a distribution over target pseudo-centroids according to Eqn. \ref{softmax}, and obtain ``soft" pseudo-labels
\STATE \textbf{\emph{Stage-3:}}  // \textbf{cycle label-consistency}
\STATE Reinforce the consistency $\mathcal{L}_{cyc}$ between ground-truth labels and pseudo-labels of source samples according to Eqn. \ref{cycle}
\STATE Compute softmax loss $\mathcal{L}_{C}$ on source domain according to Eqn. \ref{softmaxloss}
\STATE $\mathcal{L} \leftarrow \mathcal{L}_{C}+\alpha\mathcal{L}_{cyc}$
\end{algorithmic}
\end{algorithm}

In our CLCN, taking the entire training set into account and averaging the features of every class in each iteration are inefficient even impractical. Therefore, we always use mini-batch SGD for optimization in practice. However, small batch size will lead to the huge deviation between the true centroid and local centroid (centroid of current-batch samples) caused by few mislabelled samples.
%However, this strategy suffers two limitations: (1) due to the randomly selected batch, categorical information in each batch is usually incomplete. For example, it is possible that some source classes are missing in the current batch so that target data of these classes can not be correctly labeled using cross-domain nearest centroid classification algorithm. (2) Small batch size will lead to the huge deviation between the true centroid and local centroid (centroid of current-batch samples) caused by few mislabelled samples.
%To overcome the aforementioned problems, on the one hand, we suggest to appropriately set batch size to be larger to avoid incomplete categorical information; on the other hand,
To overcome this problem, global centroid \cite{chen2018progressive,xie2018learning} is utilized to approach true centroid, so that more reliable pseudo-labels are obtained through our cross-domain nearest centroid classification. In each iteration, we first compute local centroid $c_{k(t)}^s$ for source domain using the mini-batch samples according to Eqn. \ref{center}, then update the global centroid $c_{k}^s$ as follows:
\begin{equation}
c_{k}^s= \theta c_{k}^s+(1-\theta)c_{k(t)}^s \label{centroid}
\end{equation}
where $\theta$ is the trade-off parameters. The global centroid $c_{k}^t$ of target samples can be computed in the same way. %However, when computing target centroids with these more reliable target pseudo-labels, we want these centroids to represent latent structure of target samples in current batch to obtain mini-batch gradient for cycle label-consistent loss and align the cross-domain category distributions of these target samples and source samples. Therefore, we directly use local centroid for target domain.

\section{Experiments}

\begin{figure*}[htbp]
\centering
\includegraphics[width=12cm]{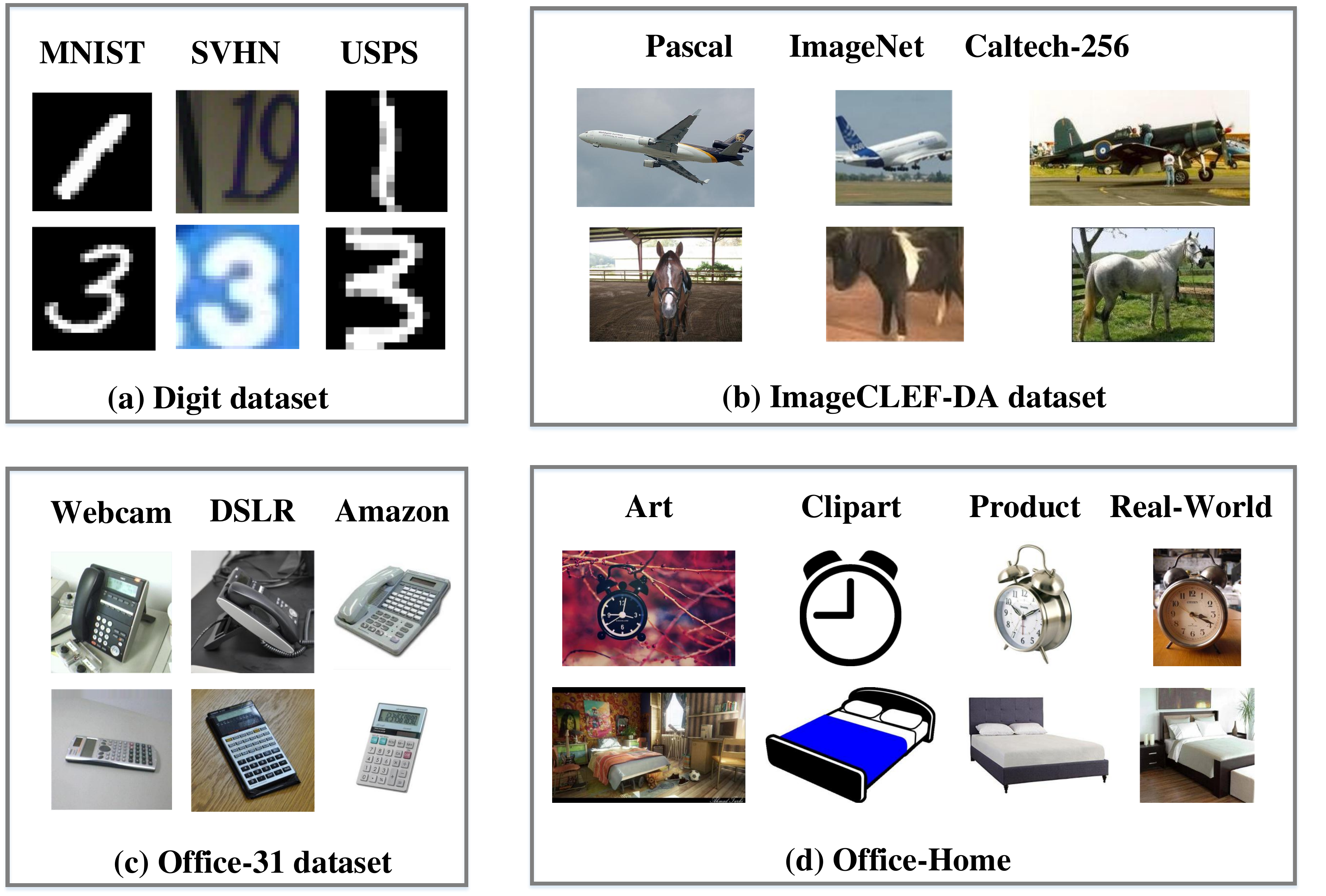}
\caption{ Dataset samples of (a) MNIST-SVHN-USPS \cite{lecun1998gradient,netzer2011reading,denker1989neural}, (b) ImageCLEF-DA \cite{caputo2014imageclef}, (c) Office-31 \cite{saenko2010adapting} and (d) Office-Home \cite{venkateswara2017deep}. }
\label{dataset}
\end{figure*}

In this section, we provide details about our implementation and training protocol, and report our experimental evaluation. Our algorithm is evaluated on various unsupervised domain adaptation tasks which focus on two different problems: hand-written digit classification and object recognition.

\subsection{Datasets}

\textbf{MNIST-SVHN-USPS.} We explore three digit datasets of varying difficulty as shown in Fig. \ref{dataset}: MNIST \cite{lecun1998gradient}, SVHN \cite{netzer2011reading} and USPS \cite{denker1989neural}. They contain digital images of 10 classes. MNIST and SVHN consist of grey images of size 28$\times$28 and 16$\times$16, respectively; USPS consists of color images of size 32$\times$32, and has extra confusing digits around the centered digit of interest. Following previous works \cite{Tzeng2017Adversarial}, we consider the three transfer tasks: MNIST$\rightarrow$SVHN, SVHN$\rightarrow$MNIST and MNIST$\rightarrow$USPS. Digit images are cast to 28$\times$28$\times$1 in all experiments for fair comparison. 

\textbf{Office-31} \cite{saenko2010adapting} is a standard dataset used for domain adaptation which contains 4110 images of 31 categories in total. This dataset contains three distinct
domains, including Amazon (A) comprising 2817 images downloaded from online merchants, Webcam (W) involving 795 low resolution images acquired from webcams, and DSLR (D) containing 498 high resolution images of digital SLRs. We evaluate our method across all 6 transfer tasks. Data augmentation such as random flipping and cropping is used in training following JAN \cite{long2017deep}.

\textbf{ImageCLEF-DA} \cite{caputo2014imageclef} is built for the ImageCLEF 2014 domain adaptation challenge with three domains, including Caltech-256 (C), ImageNet ILSVRC 2012 (I) and Pascal VOC 2012 (P). Each domain contains 12 classes and 50 images per class, which results in total 600 images for one domain. We also use data augmentation in training.

\textbf{Office-Home} \cite{venkateswara2017deep} is a better organized but more difficult dataset than Office-31, which contains 15,500 images of 65 categories in total, forming 4 domains. Specifically, Art (Ar) denotes artistic depictions for object images, Clipart (Cl) means picture collection of clipart, Product (Pr) shows object images with a clear background and is similar to Amazon category in Office-31, and Real-World (Rw) represents object images collected with a regular camera. We use all domain combinations and build 12 transfer tasks.

\subsection{Implementation detail} \label{implementation}

\textbf{CNN architecture.} For digit classification datasets, we use the same architecture with ADDA \cite{Tzeng2017Adversarial}: two convolution layers are followed by max pool layers and two fully connected layers are placed behind. Batch Normalization is inserted in convolutional layers. In experiments on Office-31, ImageCLEF-DA and Office-Home, we employ the AlexNet \cite{krizhevsky2012imagenet} pretrained on ImageNet \cite{deng2009imagenet} as our backbone networks. Following RTN \cite{long2016unsupervised} and RevGrad \cite{ganin2014unsupervised}, a bottleneck layer $fcb$ with 256 units is added after the $fc7$ layer for safer transfer representation learning. We finetune the $conv1$, $conv2$, $conv3$, $conv4$, $conv5$, $fc6$, $fc7$ layers, and train the $fcb$ layer and last classifier from scratch.

\textbf{Experimental setup.} All the experiments were implemented using Tensorflow \cite{abadi2016tensorflow}. We use mini-batch stochastic gradient descent (SGD) with momentum of 0.9 to train the network and set weight decay to $5e-4$. We set the learning rate to 0.01 on most adaptation tasks except for MNIST$\rightarrow$USPS, C$\rightarrow$P and P$\rightarrow$C adaptation tasks. The learning rate is set to be 0.001 on MNIST$\rightarrow$USPS task, and is annealed by $\mu_p=\frac{\mu_0}{\left ( 1-\gamma *p \right )^\beta } $ on C$\rightarrow$P and P$\rightarrow$C tasks, where $\mu_0=0.01$, $\gamma=10$ and $\beta=0.75$. In digit classification experiments, we set the batch size to 128. Following \cite{ranjan2017l2}, we normalize the embedding features by l2 normalization and then re-scale them to 5. In experiments on Office-31, Office-Home and ImageCLEF-DA databases, we set the batch size to 400, 400 and 128, respectively. We also normalize the embedding features by l2 normalization and then re-scale them to 10.

\textbf{Hyper-parameters.} The hyper-parameter $\theta$ is set to be 0.7 in our method. In order to suppress noisy signal at the early stages of the training procedure, we change the parameter $\alpha$ from 0 to $\alpha_0$ using the following strategy: $\alpha=\alpha_0*\left (\frac{2}{1+exp\left ( -\gamma *p \right )}-1\right )$, where $\alpha_0$ and $\gamma$ are set to 2.5 and 10, and $p$ is training progress changing from 0 to 1.

\textbf{Evaluation protocols.} We follow standard evaluation protocols for unsupervised domain adaptation as \cite{Long2015Learning,long2017deep,ganin2014unsupervised,xie2018learning}. All labeled source examples and all unlabeled target examples are used. We repeat each transfer task three times, and report the mean accuracy (number of correctly classified test samples divided by the total number of test samples) as well as the standard deviation of these three results.

\subsection{Results}

\textbf{Results on MNIST-USPS-SVHN.} We compare our method with Deep Domain Confusion (DDC) \cite{Tzeng2014Deep}, Domain Separation Networks (DSN) \cite{bousmalis2016domain}, Gradient Reversal (RevGrad), Couple GAN (CoGAN) \cite{liu2016coupled}, Adversarial Discriminative Domain Adaptation (ADDA) \cite{Tzeng2017Adversarial}, Label Efficient Learning (LEL) \cite{luo2017label}, Deep Reconstruction Classification Network (DRCN) \cite{ghifary2016deep}, Domain Adversarial Adaptation neural network (DAA) \cite{jia2019domain}, Asymmetric Tri-Training (AsmTri) \cite{saito2017asymmetric}, Moving Semantic Transfer Network (MSTN) \cite{xie2018learning} and TarGAN \cite{lv2019targan}. Table \ref{tab2} shows the performance comparisons on three transfer directions among digit datasets. 

\begin{table*}[htbp]
\caption{Classification accuracies (\%) on digit recognitions tasks.}
	\begin{center}
	\begin{tabular}{l|ccc}
		\hline
         Methods & SVHN to MNIST &  USPS to MNIST  & MNIST to USPS \\ \hline \hline
         Source-only & 60.1$\pm$1.1 & 75.2$\pm$1.6 & 57.1$\pm$1.7  \\
         DDC \cite{Tzeng2014Deep}& 68.1$\pm$0.3 & 79.1$\pm$0.5 & 66.5$\pm$3.3 \\
         DSN w/ MMD \cite{bousmalis2016domain} & 72.2 & - & - \\
         RevGrad \cite{ganin2014unsupervised} & 73.9 & 77.1$\pm$1.8 & 73.0$\pm$2.0 \\
         CoGAN \cite{liu2016coupled}  & - & 91.2$\pm$0.8 & - \\
         ADDA \cite{Tzeng2017Adversarial} & 76.0$\pm$1.8 & 89.4$\pm$0.2 & 90.1$\pm$0.8 \\
         DSN w/ DANN \cite{bousmalis2016domain} & 82.7 & - & - \\
         LEL \cite{luo2017label} & 81.0$\pm$0.3 & - & - \\
         DRCN \cite{ghifary2016deep} & 82.0$\pm$0.1 & 91.8$\pm$0.09 & 73.7$\pm$0.04 \\
         DAA \cite{jia2019domain} & 78.3$\pm$0.5 & 92.8$\pm$1.1 & 90.3$\pm$0.2 \\
         AsmTri \cite{saito2017asymmetric} & 86.0 & - & - \\
         %CyCADA \cite{hoffman2017cycada} & 90.4$\pm$0.4 & 95.6$\pm$0.2 & \textbf{96.5$\pm$0.1} \\
         MSTN \cite{xie2018learning} & 91.7$\pm$1.5 & 92.9$\pm$1.1 & - \\ 
         TarGAN \cite{lv2019targan} & \textbf{98.1} & 94.1 & 93.8 \\ \hline
         CLCN (ours) &  97.5$\pm$0.1 &  \textbf{98.5$\pm$0.1} &  \textbf{94.4$\pm$0.3}  \\
         \hline \hline
	\end{tabular}
    \end{center}
    \label{tab2}
\end{table*}

\begin{table*}[htbp]
 \caption{Classification accuracies (\%) on office-31 datasets. (AlexNet)}
	\begin{center}
    \setlength{\tabcolsep}{2mm}{
	\begin{tabular}{l|ccccccc}
		\hline
         Methods & A to W & D to W & W to D & A to D & D to A & W to A & Average \\ \hline \hline
         AlexNet \cite{krizhevsky2012imagenet}& 61.6$\pm$0.5 & 95.4$\pm$0.3 & 99.0$\pm$0.2 & 63.8$\pm$0.5 & 51.1$\pm$0.6 & 49.8$\pm$0.4 & 70.1  \\
         DDC \cite{Tzeng2014Deep}& 61.8$\pm$0.4 & 95.0$\pm$0.5 & 98.5$\pm$0.4 & 64.4$\pm$0.3 & 52.1$\pm$0.6 & 52.2$\pm$0.4 & 70.6 \\
         DAN \cite{Long2015Learning} & 68.5$\pm$0.5 & 96.0$\pm$0.3 & 99.0$\pm$0.3 & 67.0$\pm$0.4 & 54.0$\pm$0.5 & 53.1$\pm$0.5 & 72.9 \\
         DRCN \cite{ghifary2016deep} & 68.7$\pm$0.3 & 96.4$\pm$0.3 & 99.0$\pm$0.2 & 66.8$\pm$0.5 & 56.0$\pm$0.5 & 54.9$\pm$0.5 & 73.6 \\
         RevGrad \cite{ganin2014unsupervised} & 73.0$\pm$0.5 & 96.4$\pm$0.3 & 99.2$\pm$0.3 & 72.3$\pm$0.3 & 53.4$\pm$0.4 & 51.2$\pm$0.5 & 74.3 \\
         RTN \cite{long2016unsupervised} & 73.3$\pm$0.3 & 96.8$\pm$0.2 & 99.6$\pm$0.1 & 71.0$\pm$0.2 & 50.5$\pm$0.3 & 51.0$\pm$0.1 & 73.7 \\
         JAN \cite{long2017deep} & 74.9$\pm$0.3 & 96.6$\pm$0.2 & 99.5$\pm$0.2 & 71.8$\pm$0.2 & 58.3$\pm$0.3 & 55.0$\pm$0.4 & 76.0 \\
         AutoDIAL \cite{cariucci2017autodial} & 75.5 & 96.6 & 99.5 & 73.6 & 58.1 & 59.4 & 77.1 \\
         DAA \cite{jia2019domain} & 74.3$\pm$0.3 & 97.1$\pm$0.2 & 99.6$\pm$0.2 & 72.5$\pm$0.2 & 52.5$\pm$0.3 & 53.2$\pm$0.1 & 74.8 \\
         PACET \cite{liang2019exploring}& 72.2 & 96.0 & 99.4 & 70.3 & 61.8 & 59.0 & 76.5 \\
         CAADA \cite{rahman2019correlation} & 80.2 & 97.1 & 99.2 &  \textbf{77.7} & 58.1 & 57.4 & 78.3 \\
         MSTN \cite{xie2018learning} & \textbf{80.5$\pm$0.4} & 96.9$\pm$0.1 & 99.9$\pm$0.1 &74.5$\pm$0.4 & 62.5$\pm$0.4 & 60.0$\pm$0.6 & 79.1 \\  \hline
         CLCN (ours) &  78.4$\pm$0.4 &  \textbf{97.6$\pm$0.1} &  \textbf{99.9$\pm$0.2}  & 73.9$\pm$0.2 & \textbf{64.3$\pm$0.2} & \textbf{62.8$\pm$0.1} & \textbf{79.5} \\ \hline
         \hline
	\end{tabular}}
    \end{center}
    \label{tab3}
\end{table*}

From the results, we can see several important observations. 1) The performances of source-only model which is trained using only labeled source data could be regarded as a lower bound without domain adaptation. Source-only model gives the accuracies of 60.1\%, 75.2\% and 57.1\% on the SVHN$\rightarrow$MNIST, USPS$\rightarrow$MNIST and MNIST$\rightarrow$USPS adaptation tasks, respectively, showing deep network indeed suffers from the domain shift problem. 2) When matching global distribution of source and target domains through MMD or adversarial learning, DDC \cite{Tzeng2014Deep}, RevGrad \cite{ganin2014unsupervised} and ADDA \cite{Tzeng2017Adversarial} outperform the source-only model on target domain, but this improvement is limited. Because they are category agnostic leading to mismatch of label spaces across different domains. 3) When category information is incorporated, e.g. MSTN \cite{xie2018learning}, the improvement becomes more significant. By aligning centroids of source and target classes, features in same class but different domains are mapped nearby, resulting in an improved target classification accuracy. 4) Our proposed CLCN achieves superior performances against other techniques. The accuracy of CLCN can achieve 97.5\%, 98.5\% and 94.4\% on the SVHN$\rightarrow$MNIST, USPS$\rightarrow$MNIST and MNIST$\rightarrow$USPS adaptation tasks, respectively, making the absolute improvement over MSTN \cite{xie2018learning} by 5.8\% on SVHN$\rightarrow$MNIST setting and 5.6\% on MNIST$\rightarrow$USPS setting. Compared with global alignment methods, CLCN takes consideration of category information leading to more similar latent representations between two domains; while compared with centroid alignment techniques, CLCN is supervised with ground-truth labels and takes full advantage of backpropagation information provided by each sample. 

\textbf{Results on Office-31.} We also compare our method with DDC \cite{Tzeng2014Deep}, DRCN \cite{ghifary2016deep}, RevGrad \cite{ganin2014unsupervised}, DAA \cite{jia2019domain}, MSTN \cite{xie2018learning}, Deep Adaptation Networks (DAN) \cite{Long2015Learning}, Residual Transfer Network (RTN) \cite{long2016unsupervised}, Joint Adaptation Network (JAN) \cite{long2017deep}, Automatic Domain Alignment Layer (AutoDIAL) \cite{cariucci2017autodial}, Progressive leArning with Confidence-wEighted Targets (PACET) \cite{liang2019exploring} and Correlation-aware adversarial domain adaptation (CAADA) \cite{rahman2019correlation} on Office-31 dataset. The results are reported in Table \ref{tab3}.

\begin{table*}[htbp]
\caption{Classification accuracies (\%) on ImageCLEF-DA datasets. (AlexNet)}
	\begin{center}
    \setlength{\tabcolsep}{2mm}{
	\begin{tabular}{l|ccccccc}
		\hline
         Methods & I to P & P to I & I to C & C to I & C to P & P to C & Average \\ \hline \hline
         AlexNet \cite{krizhevsky2012imagenet}& 66.2$\pm$0.2 & 70.0$\pm$0.2 & 84.3$\pm$0.2 & 71.3$\pm$0.4 & 59.3$\pm$0.5 &  84.5$\pm$0.3 & 73.9  \\
         RTN \cite{long2016unsupervised}& 67.4$\pm$0.3 & 81.3$\pm$0.3 & 89.5$\pm$0.4 & 78.0$\pm$0.2 & 62.0$\pm$0.2&  89.1$\pm$0.1&  77.9 \\
         RevGrad \cite{ganin2014unsupervised} & 66.5$\pm$0.5 & 81.8$\pm$0.4 & 89.0$\pm$0.5 & 79.8$\pm$0.5 & 63.5$\pm$0.4 & 88.7$\pm$0.4 & 78.2 \\
         JAN \cite{long2017deep} & 67.2$\pm$0.5 & 82.8$\pm$0.4 & 91.3$\pm$0.5 & 80.0$\pm$0.5 & 63.5$\pm$0.4 & 91.0$\pm$0.4 & 79.3 \\
         MSTN \cite{xie2018learning} & 67.3$\pm$0.3 & 82.8$\pm$0.2 & 91.5$\pm$0.1 & \textbf{81.7$\pm$0.3} & \textbf{65.3$\pm$0.2} & 91.2$\pm$0.2 & 80.0 \\ 
         CAADA \cite{rahman2019correlation} & 67.8 & \textbf{84.5} & 91.7 & 81.3 & 63.9 & 91.8 & 80.2 \\ \hline
         CLCN (ours) &  \textbf{68.5$\pm$0.2} & 84.2$\pm$0.6 & \textbf{91.8$\pm$0.4} & 79.9$\pm$0.1 & 64.7$\pm$0.2 & \textbf{92.2$\pm$0.2} & \textbf{80.2} \\ \hline
         \hline
	\end{tabular}}
    \end{center}
        \label{tab4}
\end{table*}

\begin{table*}[htbp]
\caption{Classification accuracies (\%) on Office-Home datasets. (AlexNet)}
	\begin{center}
    \setlength{\tabcolsep}{2mm}{
	\begin{tabular}{l|ccccccccccccc}
		\hline
         Source & Ar & Ar & Ar & Cl & Cl & Cl & Pr & Pr & Pr & Rw & Rw & Rw & \multirow{2}{*}{Avg} \\
         Target   & Cl & Pr & Rw & Ar & Pr & Rw & Ar & Cl & Rw & Ar &Cl  & Pr & \\ \hline \hline
         GFK \cite{gong2012geodesic}    & 21.60 & 31.72 & 38.83 & 21.63 & 34.94 & 34.20 & 24.52 & 25.73  & 42.92 & 32.88 & 28.96 & 50.89 & 32.40 \\
         JDA \cite{long2013transfer}        & 25.34 & 35.98 & 42.94 & 24.52 & 40.19 & 40.90&  25.96 & 32.72 & 49.25 & 35.10 & 35.35 & 55.35  & 36.97 \\
         CCSL \cite{ming2015unsupervised}   & 23.51 & 34.12 & 40.02 & 22.54 & 35.69 & 36.04 & 24.84 & 27.09 & 46.36 & 34.61 & 31.75 & 52.89 &  34.12 \\
         LSC \cite{hou2016unsupervised}                     & 31.81 & 39.42 & 50.25 & 35.46 & 51.19 & 51.43 & 30.46 & 39.54 & 59.74 & 43.98 & 42.88 & 62.25 & 44.87 \\
         RTML \cite{ding2016robust}                  & 27.57 & 36.20 & 46.09 & 29.49 & 44.69 & 44.66 & 28.21 & 36.12 & 52.99 & 38.54 & 40.62 & 57.80 & 40.25 \\
         JGSA \cite{zhang2017joint}                  & 28.81 & 37.57  & 48.92 & 31.67 & 46.30 & 46.76 & 28.72 & 35.90 & 54.47 & 40.61 & 40.83 & 59.16 & 41.64 \\
         PUnDA \cite{gholami2017punda}               & 29.99 & 37.76 & 50.17 & 33.90 & 48.91 & 48.71 & 30.31 & 38.69 & 56.91 & 42.25 & 44.51 & 61.05 & 43.60 \\ \hline
         DAN \cite{Long2015Learning}                   & 30.66 & 42.17 & 54.13 & 32.83 & 47.59 & 49.78 & 29.07 & 34.05 & 56.70 & 43.58 & 38.25 & 62.73 & 43.46 \\
         DHN \cite{venkateswara2017deep}                   & 31.64 & 40.75 & 51.73 & 34.69 & 51.93 & 52.79 & 29.91 & 39.63 & 60.71 & 44.99 & 45.13 & 62.54 & 45.54 \\
         WDAN \cite{yan2017mind}             & 32.26  & 43.16 & 54.98 & 34.28 & 49.92 & 50.26 & 30.82 & 38.27 & 56.87 & 44.32 & 39.35 & 63.34 &44.82 \\
         GAKT \cite{ding2018graph}              & 34.49 & 43.63 & 55.28 & 36.14 & 52.74 & 53.16 & 31.59 & 40.55 & \textbf{61.43} & 45.64 & 44.58 & 64.92 &  47.01 \\
         MSTN \cite{xie2018learning}             & 34.87 & 46.20 & 56.77 & 36.63 & 54.97 & 55.41 & 33.27 &  \textbf{41.66} & 60.62 & 46.94 & 45.90 & 68.25 & 48.46 \\ \hline
         \multirow{2}{*}{CLCN (ours)} &  \textbf{37.58} & \textbf{49.39} & \textbf{57.70} & \textbf{37.08} & \textbf{55.31} & \textbf{56.24} & \textbf{34.80} & 39.85 & 61.03 & \textbf{48.63} & \textbf{46.14} & \textbf{68.93} & \multirow{2}{*}{\textbf{49.39}} \\
          & \textbf{$\pm$0.2} & \textbf{$\pm$0.3} & \textbf{$\pm$0.2} & \textbf{$\pm$0.1} & \textbf{$\pm$0.4} & \textbf{$\pm$0.1} & \textbf{$\pm$0.1} & $\pm$0.4 & $\pm$0.2 & \textbf{$\pm$0.1} & \textbf{$\pm$0.2} & \textbf{$\pm$0.1} & \\ \hline
         \hline
	\end{tabular}}
    \end{center}
    \label{tab6}
\end{table*}

Observing the results in Table \ref{tab3}: 1) without adaptation, AlexNet \cite{krizhevsky2012imagenet} can not obtain perfect performance on target domain due to domain shift. 2) Existing domain-level alignment methods, such as RTN \cite{long2016unsupervised} and RevGrad \cite{ganin2014unsupervised}, help to boost performance by reducing this discrepancy. 3) Our approach outperforms comparison methods on most adaptation tasks, which reveals that cycle label-consistency is effective and CLCN is scalable for different datasets. For example, we observe that CLCN is superior to JAN \cite{long2017deep} significantly by about 3\%. Compared with the best competitor, i.e. MSTN \cite{xie2018learning}, our CLCN obtains comparable results by a simpler and reliable way. MSTN performs domain-level alignment by adversarial learning and performs class-level alignment by minimizing the distances between source and target centroids with the same labels; while our CLCN only depends on a single penalty item, i.e. cycle label-consistent loss. The results demonstrate the effectiveness of our method which reinforces the consistency between ground-truth labels and pseudo-labels of source samples. %It optimizes networks based on backpropagation information provided by samples and is supervised with ground-truth labels, resulting in statistically similar latent representations between source and target domain and compact latent space clustering.

\textbf{Results on ImageCLEF-DA.} We compare our method with DAN \cite{Long2015Learning}, RTN \cite{long2016unsupervised}, RevGrad \cite{ganin2014unsupervised}, JAN \cite{long2017deep}, MSTN \cite{xie2018learning} and CAADA \cite{rahman2019correlation} on ImageCLEF-DA dataset. The results are reported in Table \ref{tab4}. The improvement on ImageCLEF-DA is less than Office-31 since the difference in domain sizes will cause more serious shift \cite{long2017deep}. The improvement of CLCN over Alexnet w.r.t. the average accuracy is 6.3\%. Among the counterparts using domain adaptation algorithms, CAADA is the state-of-the-art approach. Our CLCN ranks within top two in both 5 out of 6 tasks, and it outperforms CAADA in most tasks.
These convincing results on the challenging ImageCLEF-DA dataset indicate that our method has the potential to generalize to a variety of settings.

\textbf{Results on Office-Home.} On Office-Home dataset, we compare our CLCN with some shallow methods, e,g, Geodesic Flow Kernel (GFK) \cite{gong2012geodesic}, Joint Geometrical and Statistical Alignment (JGSA) \cite{zhang2017joint}, Probabilistic Unsupervised Domain Adaptation (PUnDA) \cite{gholami2017punda}, as well as deep methods, e.g. DAN \cite{Long2015Learning}, MSTN \cite{xie2018learning}, Deep Hashing Network (DHN) \cite{venkateswara2017deep}, Weighted Domain Adaptation Network (WDAN) \cite{yan2017mind}, Graph Adaptive Knowledge Transfer (GAKT) \cite{ding2018graph}. The results of Office-Home are reported in Table \ref{tab6}.  

\begin{figure*}
\centering
\subfigure[Non-adapted]{
\label{fig3a} %% label for first subfigure
\includegraphics[width=4.3cm]{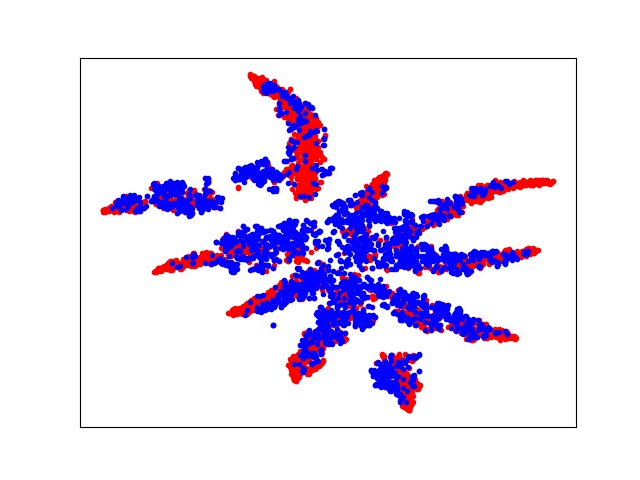}}
\subfigure[Adversarial Adapted]{
\label{fig3b} %% label for second subfigure
\includegraphics[width=4.3cm]{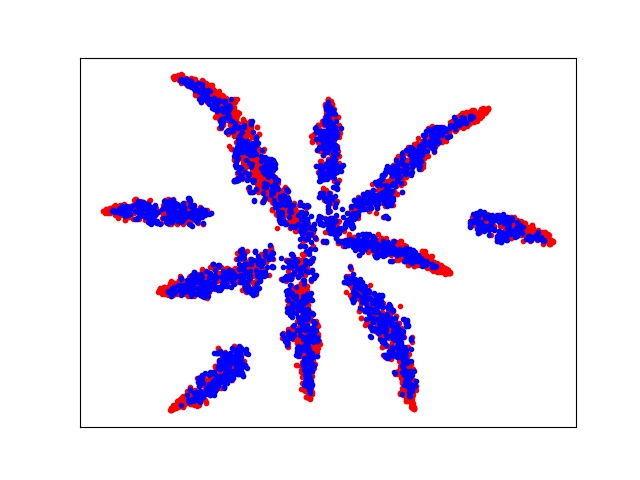}}
\subfigure[Cycle Adapted]{
\label{fig3c} %% label for first subfigure
\includegraphics[width=4.3cm]{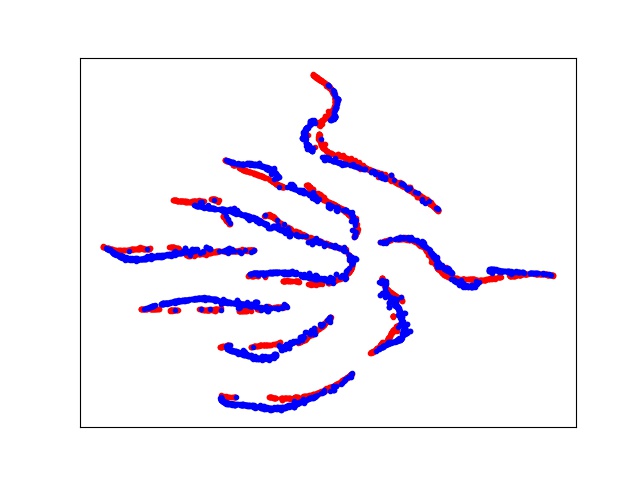}}
\caption{Feature visualization on the SVHN$\rightarrow$MNIST task. We confirm the effects of CLCN through a visualization of the learned representations using t-distributed stochastic neighbor embedding (t-SNE) \cite{maaten2008visualizing}. Red points are source samples and Blue are target samples. (a) is trained without any adaptation, (b) is trained with previous adversarial domain adaptation method, i.e. RevGrad \cite{ganin2014unsupervised}, (c) is trained with our CLCN method. As we can see, compared to non-adapted method, the features generated by RevGrad \cite{ganin2014unsupervised} are successfully fused but are not discriminated and compact. The features near class boundary are obviously harmful to classification tasks. Our CLCN method aligns feature space and forms compact clusters leading to an improved performance.}
\label{fig3} %% label for entire figure
\end{figure*}

\begin{figure*}[htbp]
\centering
\subfigure[Accuracy SVHN$\rightarrow$MNIST]{
\label{fig4a} %% label for first subfigure
\includegraphics[width=4.2cm]{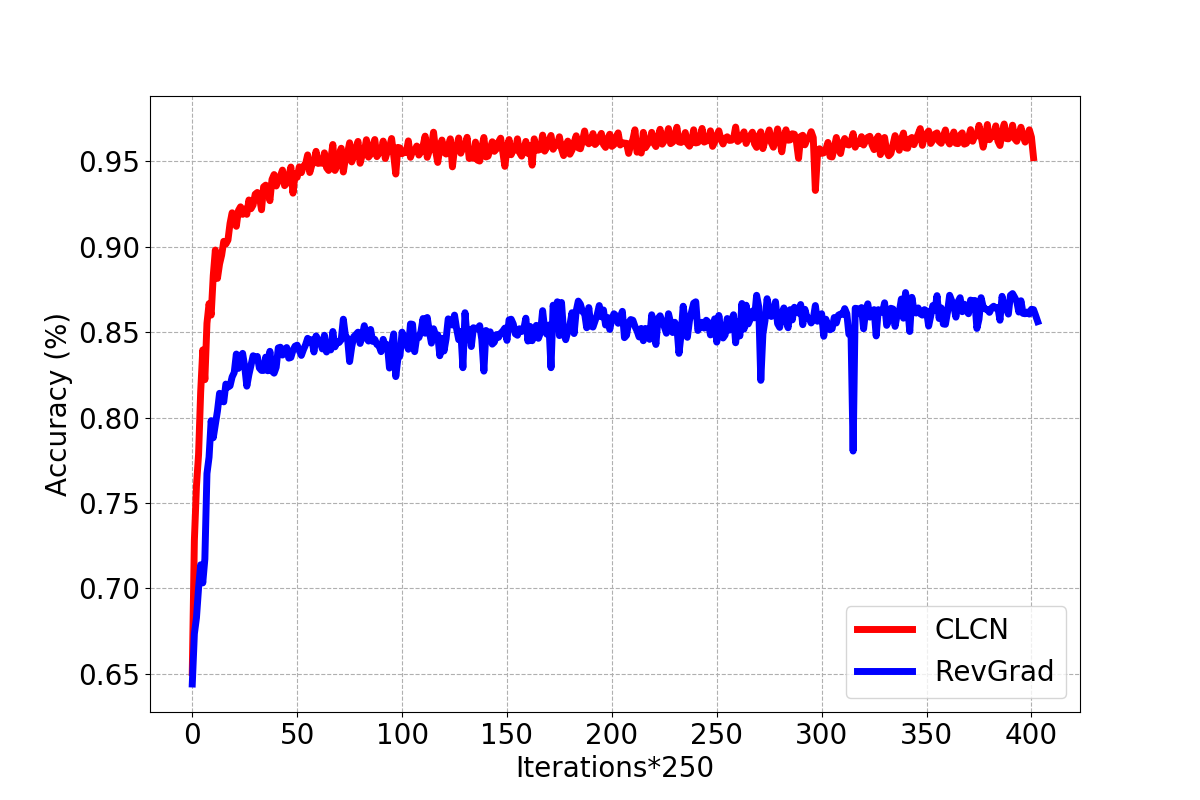}}
\hspace{0cm}
\subfigure[Distance SVHN$\rightarrow$MNIST]{
\label{fig4b} %% label for second subfigure
\includegraphics[width=4.2cm]{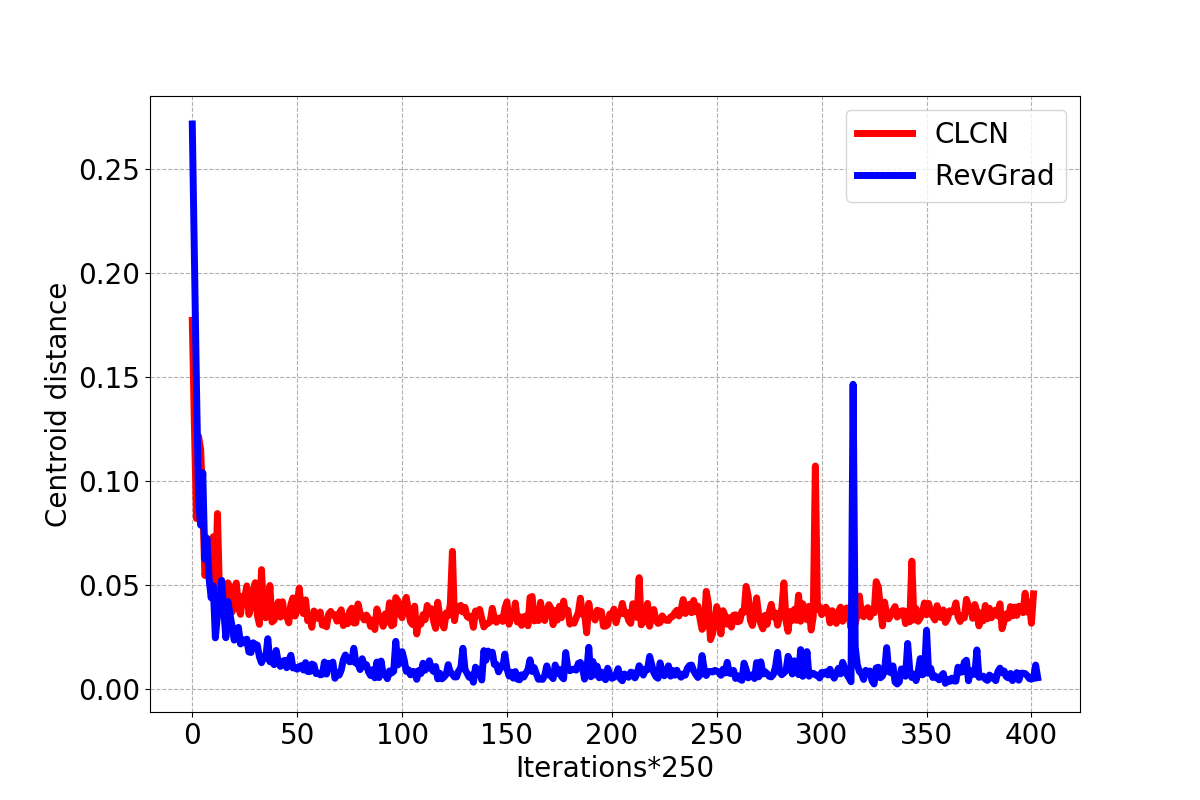}}
\hspace{0cm}
\subfigure[Accuracy D$\rightarrow$A]{
\label{fig4c} %% label for first subfigure
\includegraphics[width=4.2cm]{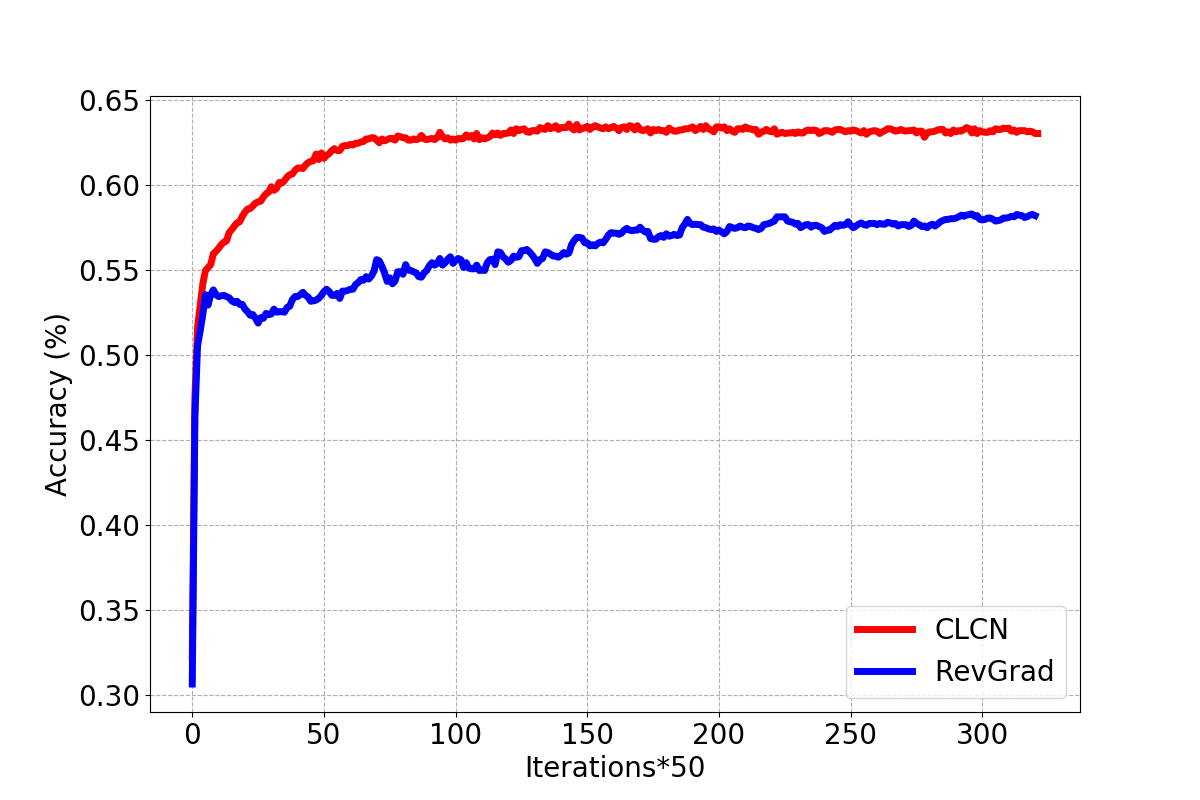}}
\hspace{0cm}
\subfigure[Distance D$\rightarrow$A]{
\label{fig4d} %% label for first subfigure
\includegraphics[width=4.2cm]{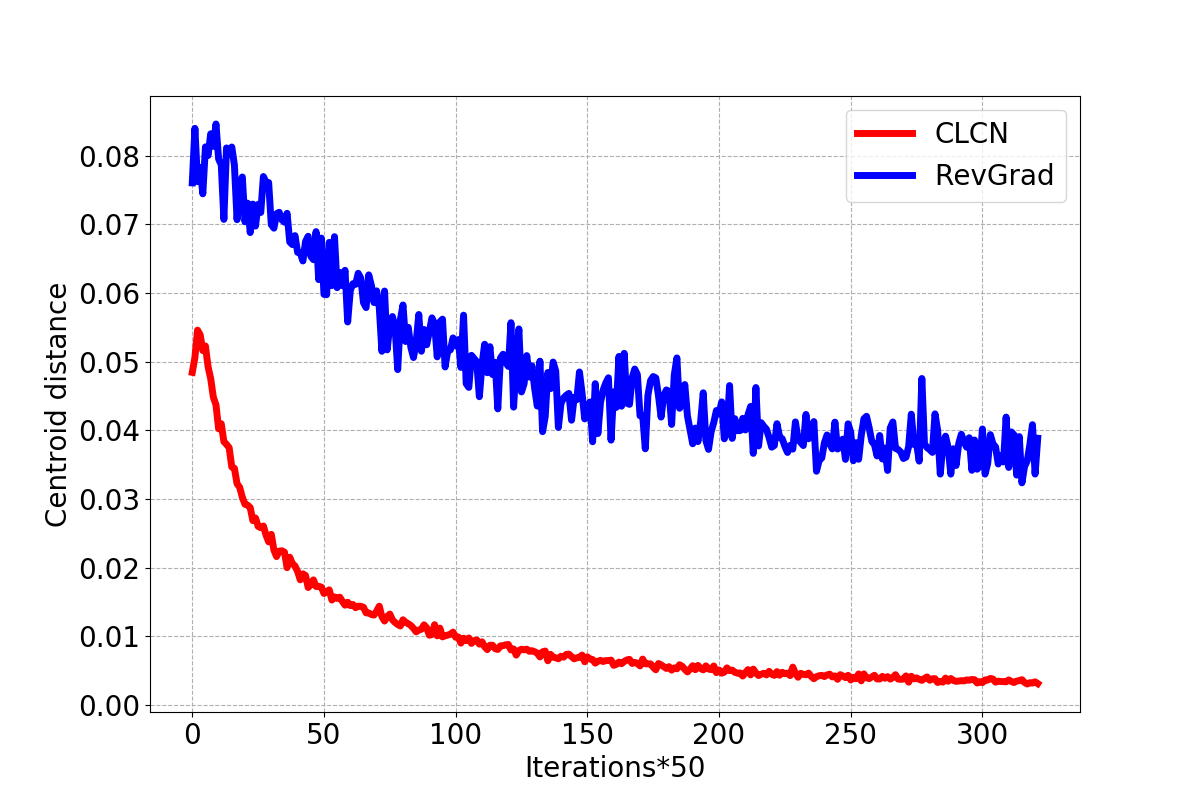}}
\caption{RevGrad \cite{ganin2014unsupervised} in blue, our model CLCN in red. (a)(c): Comparison of testing accuracies of RevGrad and our proposed method CLCN on SVHN$\rightarrow$MNIST and D$\rightarrow$A adaptation task. Our model has similar convergence speed as RevGrad. (b)(d): The distances between source centroids and target centroids with the same labels. For the easier digit tasks, such as SVHN$\rightarrow$MNIST, the centroid distances of the same classes can be both reduced to a small value in RevGrad and CLCN model; while for the harder tasks, such as D$\rightarrow$A, the centroid distances in CLCN model are much smaller than those in RevGrad model.}
\label{fig4} %% label for entire figure
\end{figure*}

We can see that our CLCN outperforms the comparison methods on most transfer tasks, and exceeds the MSTN \cite{xie2018learning} about 1\% by average. And we have the following observations. 1) Global alignment methods, i.e. DAN \cite{Long2015Learning}, can only obtain limited improvement on Office-Home datasets. Although benefiting from better representations of deep learning, deep methods, e.g. DAN \cite{Long2015Learning}, obtain a similar level of performance compared with shallow methods, e.g. PUnDA \cite{gholami2017punda}. The reason may be that the four domains in Office-Home are with more categories, and are visually more dissimilar with each other. Since domain alignment is category agnostic in previous work, it is possible that the aligned domains are not classification friendly in the presence of large number of categories. This confirms that global alignment is not enough, and other constraints, e.g. compact clusters in latent space, are vital in domain adaptation problem. 2) Our CLCN yields larger boosts on Office-Home domain adaptation tasks. It takes into account the cycle consistency of classification across domain, resulting in higher similarity within the same class and better tightness of the clusters.

\subsection{Experimental analysis} \label{analysis}

\textbf{Feature visualization.} To demonstrate the transferability of the CLCN learned features, the visualization comparisons are conducted at feature level. First, we randomly extract the deep features of source and target images in the SVHN$\rightarrow$MNIST task with source-only model, RevGrad \cite{ganin2014unsupervised} model and CLCN model, respectively. The features are visualized using t-distributed stochastic neighbor embedding (t-SNE) \cite{maaten2008visualizing}, as shown in Fig. \ref{fig3}. Fig. \ref{fig3a} shows the representations without any adaptation. As we can see, the distributions are separated between domains, which visually proves that there is domain shift between images of SVHN and those of MNIST. Fig. \ref{fig3b} shows the result of RevGrad method. Although features are successfully fused, the target points are not discriminated and compact. The ambiguous features are generated near class boundary which are obviously harmful to classification tasks. Fig. \ref{fig3c} shows the representations of our CLCN method. We can see that the features with the same labels are concentrated and form tight clusters, and those from different classes are separated. Therefore, we conclude that the cycle label-consistency does help our CLCN to align feature space and form compact clusters at class level so that the target presentations are more discriminated and the performance of target domain is improved.

\textbf{Convergence analysis.} To inspect how CLCN converges, we show the test accuracy with respect to the number of iterations in Fig. \ref{fig4a} and \ref{fig4c}. On SVHN$\rightarrow$MNIST and D$\rightarrow$A adaptation tasks, CLCN shows similar convergence rate with RevGrad \cite{ganin2014unsupervised} but better performance. Moreover, to verify that our cycle label-consistency loss can learn statistically similar latent representations between two domains, we further examine distances between source centroids and target centroids with the same labels. We compute global centroid of each source class and each target class according to Eqn. \ref{centroid}. The squared Euclidean distance $d(x,y)=\|x-y\|^2$ is utilized, so the total centroid distance can be formulated as: $d(c^{s},c^{t})=\sum_{k=\hat{k}=1}^{K}\|c^{s}_{k}-c^{t}_{\hat{k}}\|^2$. The results of centroid distances with respect to the number of iterations are shown in Fig. \ref{fig4b} and \ref{fig4d}. For the SVHN$\rightarrow$MNIST adaptation task, the centroid distances can both be reduced to a small value in RevGrad \cite{ganin2014unsupervised} and CLCN model due to the simplicity of digit adaptation tasks. It is coincident with the observation in feature visualization in Fig. \ref{fig3b} and \ref{fig3c} where the source and target features are both successfully fused and aligned in RevGrad and CLCN model. However, our CLCN model is superior to RevGrad in the SVHN$\rightarrow$MNIST task since CLCN additionally encourages each source sample to cluster into corresponding centroid so that more compact and discriminative presentations are learned. For the harder tasks, such as D$\rightarrow$A, the centroid distances in CLCN model are much smaller than those in RevGrad model. The global alignment method, e.g. RevGrad model, can not effectively pass the class information to the adaptation network and can not guarantee that samples from different domains but with the same class label will map nearby in the feature space. Our CLCN model achieves stricter alignment resulting in better performance.

\begin{figure}
\centering
\includegraphics[width=7cm]{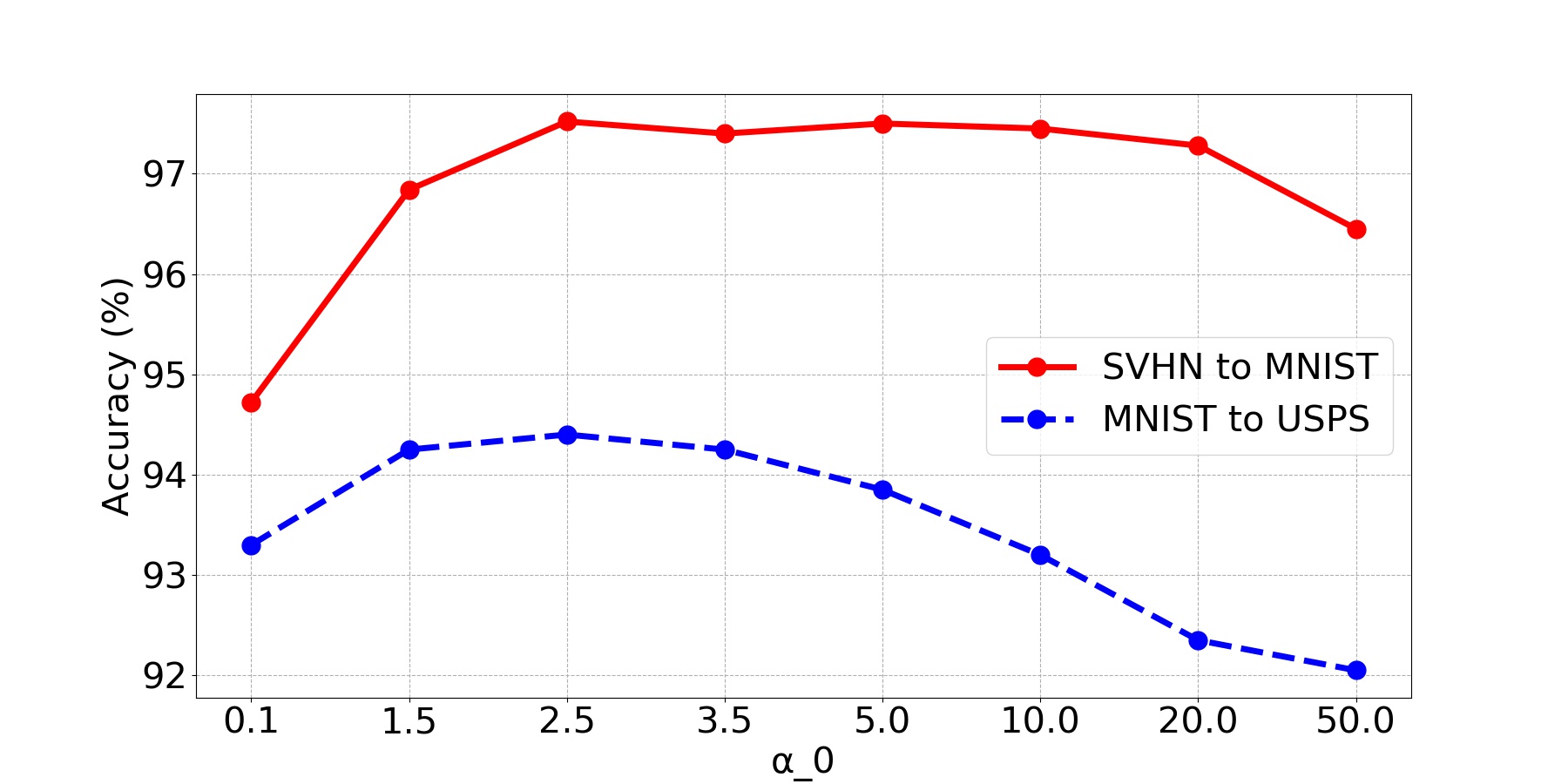}
\caption{ Parameter sensitivity study on the SVHN$\rightarrow$MMIST and MNIST$\rightarrow$USPS adaptation tasks.}
\label{sensitivity}
\end{figure}

\textbf{Parameter sensitivity.} Our CLCN is optimized by source classification loss and cycle label-consistent loss, and achieves comparable performance with other complicated domain adaptation methods. As mentioned in Section \ref{implementation}, we set the trade-off hyper-parameter $\alpha$ as $\alpha_0*\left (\frac{2}{1+exp\left ( -\gamma *p \right )}-1\right )$ for most experiments, where $\frac{2}{1+exp\left ( -\gamma *p \right )}-1$ is the commonly-used strategy \cite{xie2018learning,ganin2014unsupervised} which gradually changes adaptation factor from 0 to 1 to suppress noisy signal at the early stages of the training procedure. To have a closer look at the parameter $\alpha_0$, we perform sensitivity analysis for it on the SVHN$\rightarrow$MMIST and MNIST$\rightarrow$USPS adaptation task by varying the parameter of interest in \{0.1, 1.5, 2.5, 3.5, 5, 10, 20, 50\}, and show the result in Fig. \ref{sensitivity}. From Fig. \ref{sensitivity}, we can see that the performance of our model remains largely stable across a wide range of parameter values on SVHN$\rightarrow$MMIST tasks. And, the accuracy first increases and then decreases as $\alpha_0$ varies and demonstrates a desirable bell-shaped curve. In principle, smaller values of $\alpha_0$ could avoid CLCN learning perfect domain transfer features, and larger values of $\alpha_0$ will weaken the effects of source classification loss and degrade the classification performance.

\textbf{Ablation study.} We perform an ablation study to investigate the effectiveness of two proposed techniques (i.e., cross-domain nearest centroid classification and cycle label-consistency). Particularly, we introduce two different variants of CLCN, i.e. \emph{softmax-based cycle (w/o NCC)} and \emph{NCC-based finetuning (w/o cycle)}. In \emph{softmax-based cycle (w/o NCC)} method, NCC is replaced by two softmax classifiers which are used to generate pseudo labels for two domains. After obtaining pseudo labels, it reinforces consistency between ground-truth labels and pseudo-labels of source samples. \emph{NCC-based finetuning (w/o cycle)} method generates target pseudo-labels through source2target nearest centroid classification (NCC), and then directly utilizes these pseudo labels to finetune the network instead of performing cycle label-consistency. We show the comparison results in Table \ref{ablation}. 

\begin{table}
    \small
    \caption{Ablation study on SVHN$\rightarrow$MNIST and A$\rightarrow$W adaptation tasks.}
	\begin{center}
    \begin{threeparttable}
    \setlength{\tabcolsep}{5mm}{
	\begin{tabular}{l|cc}
		\hline
         Methods & SVHN to MMIST & A to W  \\ \hline \hline
         w/o NCC &70.6$\pm$0.2& 65.5$\pm$0.3\\
         w/o cycle & 82.2$\pm$0.3 & 72.1$\pm$0.4  \\
         CLCN & \textbf{97.5$\pm$0.1} & \textbf{78.4$\pm$0.4} \\
         %CLCN(w/o entropy) (ours) & 68.3$\pm$0.3 \\
         \hline
	\end{tabular}}
    \end{threeparttable}
    \end{center}
    \label{ablation}
\end{table}

First, our CLCN is consistently superior to \emph{NCC-based finetuning (w/o cycle)} method, which indicates the importance of such a cycle consistency technique. The \emph{NCC-based finetuning (w/o cycle)} method only obtains 82.2\% and 72.1\% on the SVHN$\rightarrow$MNIST and A$\rightarrow$W tasks, respectively. Our CLCN achieves better performances and is superior to \emph{NCC-based finetuning (w/o cycle)} by about 15.3\% on the SVHN$\rightarrow$MNIST task and 6.3\% on the A$\rightarrow$W task. Due to domain shift, the network trained with source data can not assign pseudo-labels for all target samples correctly. Some target samples lay far away from the source domain and they are ambiguous for the classification boundaries. These false-labeled samples introduce wrong information in \emph{NCC-based finetuning (w/o cycle)} method and potentially result in the error accumulation. Rather than back-propagating the category loss based on each pseudo-labeled sample, our CLCN just utilizes the centroids of target pseudo-classes to transport label information to source domain. The centroids computed by averaging the features of each class can alleviate the wrong information of several falsely-labeled samples. When label information is transported back to source domain, cycle label-consistent loss is supervised with ground-truth labels of source samples, which makes our CLCN more reliable compared with \emph{NCC-based finetuning (w/o cycle)} method.

\begin{figure*}
\centering
\subfigure[Pr$\rightarrow$Cl]{
\label{ratio_a} %% label for first subfigure
\includegraphics[width=5.8cm]{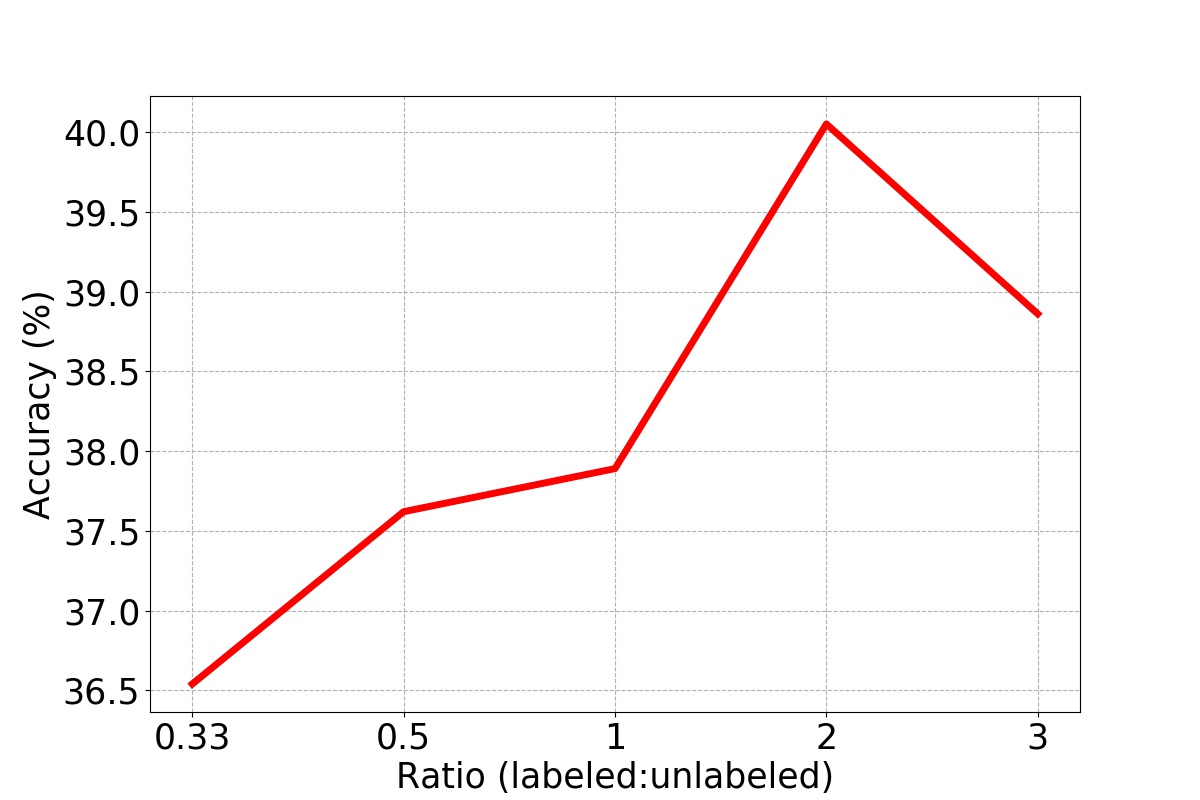}}
\hspace{0.3cm}
\subfigure[Rw$\rightarrow$Pr]{
\label{ratio_b} %% label for second subfigure
\includegraphics[width=5.8cm]{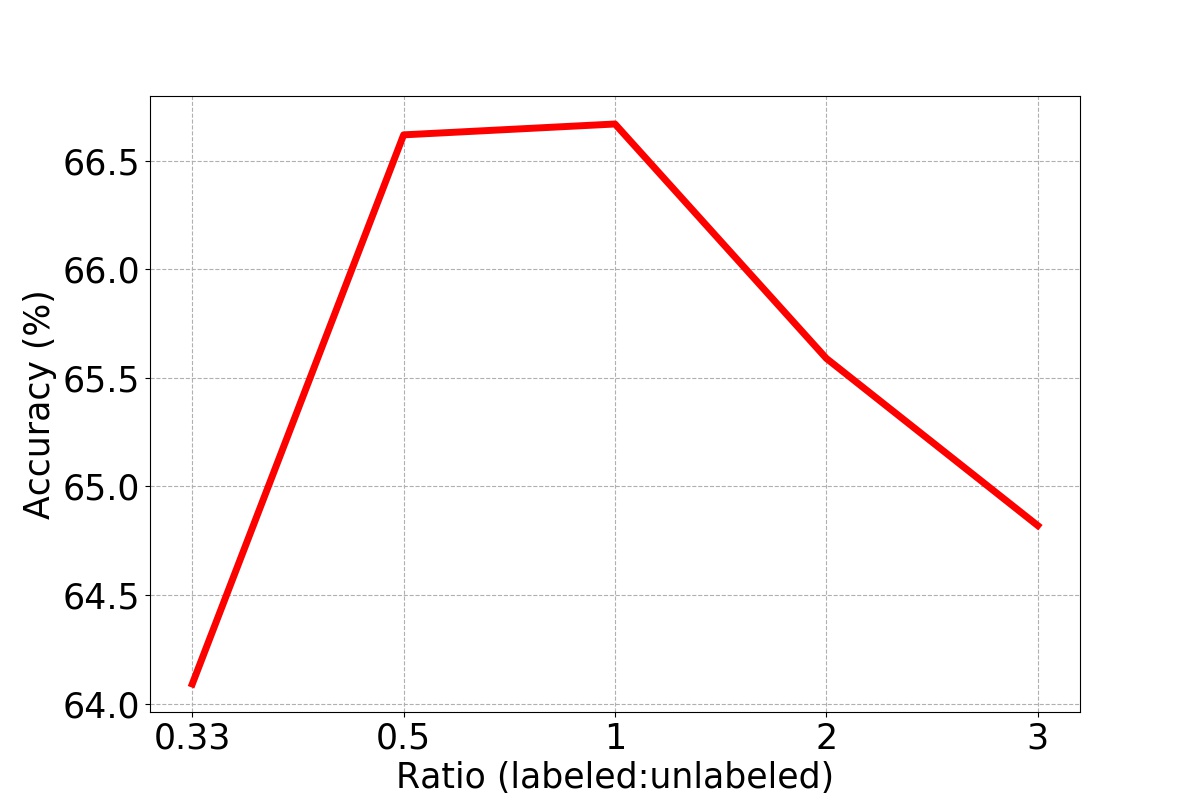}}
\caption{Accuracy on Pr$\rightarrow$Cl and Rw$\rightarrow$Pr tasks with different ratios of labeled source samples to unlabeled target samples.}
\label{ratio} %% label for entire figure
\end{figure*}

Second, compared with our CLCN, \emph{softmax-based cycle (w/o NCC)} method should additionally train a target softmax classifier to generate source pseudo labels, which is an extra burden. Moreover, we can see from the results that our CLCN outperforms \emph{softmax-based cycle (w/o NCC)} by a large margin. Our hypothesis to explain this phenomenon is that the classification ability of softmax classifier is superior to NCC algorithm which makes it easier for target softmax classifier to overfit the target pseudo-labels generated by the source classifier. When target softmax classifier generates pseudo-labels for source domain, this overfitting results in high similarity between pseudo-labels and ground-truth labels of source samples so that cycle label-consistent loss can not obtain enough gradient to optimize the network.

\begin{table}
\caption{Classification accuracies (\%) of different finetuning methods on the SVHN$\rightarrow$MNIST and A$\rightarrow$W adaptation tasks.}
    \small
	\begin{center}
    \begin{threeparttable}
    \setlength{\tabcolsep}{3mm}{
	\begin{tabular}{l|cc}
		\hline
         Methods & SVHN to MMIST & A to W  \\ \hline \hline
         AsmTri\tnote{1} \cite{saito2017asymmetric} & 86.0 & - \\
         softmax (finetuning) &86.6$\pm$2.7& 72.5$\pm$1.2\\
         NCC (finetuning) & 82.2$\pm$0.3 & 72.1$\pm$0.4  \\
         NCC (cycle+finetuning)  & \textbf{98.5$\pm$0.1} & 75.4$\pm$0.6  \\
         CLCN (ours)& 97.5$\pm$0.1 & \textbf{78.4$\pm$0.4} \\
         %CLCN(w/o entropy) (ours) & 68.3$\pm$0.3 \\
         \hline
	\end{tabular}}
    \begin{tablenotes}
     \item[1] AsmTri \cite{saito2017asymmetric} utilizes two source softmax classifiers to generate target pseudo-labels through voting, then finetunes network with them.
    \end{tablenotes}
    \end{threeparttable}
    \end{center}
    \label{tab5}
\end{table}

\textbf{Comparison with other finetuning mehtods.} To verify the effectiveness of our CLCN, we additionally compare CLCN with other finetuning methods, e.g. \emph{softmax-based finetuning}, on the SVHN$\rightarrow$MNIST and A$\rightarrow$W adaptation tasks, and show the results in Table \ref{tab5}. Although \emph{softmax-based finetuning} method performs better than \emph{NCC-based finetuning} on these two tasks, it is still worse than our CLCN. Moreover, we additionally compare our CLCN with \emph{NCC-based cycle and finetuning} method. It first generates pseudo labels by NCC, and then performs the cycle label-consistency and finetuning simultaneously. As we can see from Table \ref{tab5}, \emph{NCC-based cycle and finetuning} achieves about 1\% gain over CLCN on the SVHN$\rightarrow$MMIST adaptation task; but its result is unsatisfactory on the A$\rightarrow$W task. This phenomenon is caused by the quality of target pseudo-labels. Benefiting from our CLCN, well aligned and compact presentations are learned in the easier task SVHN$\rightarrow$ MMIST. Therefore, most target samples are correctly labeled through target2source nearest centroid classification. Finetuning network with these reliable pseudo-labels is really helpful. When performing CLCN on the harder task A$\rightarrow$W, the performance of target domain is still not perfect enough due to the large domain discrepancy. Even if finetuning method is combined with cycle label-consistency, more target samples will be wrongly labeled leading to the error accumulation and performance corrosion.

\textbf{Sensitivity to the number of samples.} In order to qualify how recognition performance changes as the number of samples is varied, we train our approach using different numbers of source and target samples on Pr$\rightarrow$Cl and Rw$\rightarrow$Pr tasks, respectively. We keep the total number of samples in two domains unchanged, and randomly select a part of data of each class from Office-home dataset. We vary the ratio of labeled source samples to unlabeled target samples ranging from $1/3$ to 3, and observe the influence of different ratios on the recognition performance. The results are shown in Fig. \ref{ratio}. We can see that the accuracy first increases and then decreases as ratio varies and demonstrates a bell-shaped curve. First, without enough labeled data, the network cannot lean powerful representations and assign pseudo-labels correctly leading to poorer performance on target domain. Second, when the number of target data is small, less label information can be transported from target domain back to the source domain which weakens the effect of our cycle label-consistency.

\section{Conclusion}

In this paper, we propose a simple yet efficient method, i.e. Cycle Label-Consistent Networks (CLCN), for unsupervised domain adaptation. We exploit cycle label-consistency and cross-domain nearest centroid classification algorithm to learn statistically similar latent representations between source and target domains and regularize the latent space to form compact clusters at class level. Especially, ``soft" pseudo-labels are generated to encourage stricter alignment and more compact clusters. Benefiting from being supervised with ground-truth lables, our CLCN can alleviate the negative influence of falsely-labeled samples without assistant of other technologies, and can take full advantage of backpropagation information provided by each sample. We experimentally show that CLCN optimized by classification loss and cycle label-consistent loss can achieve comparable performance with other complicated domain adaptation methods.

However, there are still some aspects to be improved. 1) Our method is somewhat heuristic and lacks some theoretical justification. We would like to discover some theoretical insights behind our method. 2) It highly relies on the closed set assumption where two domains share the same label space. In the future, we aim to extend it from the closed set setting to some challenging settings like open-set domain adaptation. 3) The quality of target pseudo-labels can be further improved. We consider to use the easy-to-hard scheme which progressively selects reliable pseudo-labeled target samples in the future work.

\section{Acknowledgments}

This work was partially supported by National Key R\&D Program of China (2019YFB1406504) and BUPT Excellent Ph.D. Students Foundation CX2020207.

{
\bibliographystyle{IEEEtran}
\bibliography{egbib}
}

\end{document}